\documentclass[10pt,twocolumn,letterpaper]{article}

\usepackage[pagenumbers]{cvpr} 

\usepackage{booktabs}
\usepackage{multirow}
\usepackage{graphicx}
\usepackage{colortbl}
\usepackage{xcolor}
\usepackage{amsmath}
\usepackage{amssymb}
\usepackage{adjustbox}
\definecolor{softred}{RGB}{255, 195, 195}   
\definecolor{softblue}{RGB}{185, 215, 255}  
\definecolor{hirow}{gray}{0.80}
\definecolor{hicell}{RGB}{222,235,247}
\newcommand{\best}[1]{\textbf{#1}}
\definecolor{hiloss}{RGB}{252,224,224}

\usepackage[most]{tcolorbox}
\tcbuselibrary{breakable}
\definecolor{pbSpeakF}{HTML}{2B5C8A}\definecolor{pbSpeakB}{HTML}{EDF3F9}
\definecolor{pbListF}{HTML}{4A5568} \definecolor{pbListB}{HTML}{F1F3F6}
\definecolor{pbScoreF}{HTML}{6B5B7B}\definecolor{pbScoreB}{HTML}{F4F1F6}
\definecolor{pbCentF}{HTML}{3F6B55} \definecolor{pbCentB}{HTML}{EDF4F0}
\newtcolorbox{promptbox}[2][pbSpeak]{%
  enhanced, breakable, sharp corners=downhill, arc=2pt,
  colback=#1B, colframe=#1F, boxrule=0.6pt, left=5pt, right=5pt, top=4pt, bottom=4pt,
  fonttitle=\footnotesize\bfseries\sffamily, title={#2},
  attach boxed title to top left={xshift=6pt, yshift=-2.4pt},
  boxed title style={colback=#1F, sharp corners, boxrule=0pt, left=4pt, right=4pt}}
\newcommand{\pbt}[1]{{\ttfamily\scriptsize #1}}

\definecolor{cvprblue}{rgb}{0.21,0.49,0.74}
\usepackage[pagebackref,breaklinks,colorlinks,allcolors=cvprblue]{hyperref}

\usepackage[capitalize]{cleveref}

\crefname{section}{Appendix}{Appendices}
\Crefname{section}{Appendix}{Appendices}
\def\confName{CVPR}
\def\confYear{2027}

\title{Bridging Blind Spots With Words: Decentralized Multi-View Spatial Reasoning via Language-Constrained Communication}

\author{
  Soumyabrata Chaudhuri$^{1}$\thanks{Corresponding author} \qquad
  Joydeep Ghosh$^{1}$ \\
  $^{1}$The University of Texas at Austin \\
  {\tt\small \{schaudhuri, jghosh\}@utexas.edu}
}

\begin{document}
\maketitle

\begin{abstract}
As multimodal LLMs (MLLMs) get increasingly used as collaborative partners across embodied robotics, AR/VR, and multi-user systems, spatial reasoning is now often distributed across agents with  partial environment visibility, necessitating spatial reasoning integration. Prior work in language-mediated spatial integration is however strictly limited to simplified two-agent scenarios. To address this gap, we first formalize the general problem of decentralized multi-view spatial reasoning, where N agents observe distinct viewpoints, exchange budget-constrained natural language messages, and collaborate to enable a listener to answer spatial queries. We then introduce a novel multi-view spatial benchmark 
to isolate queries that demand non-trivial multi-view integration. Through an extensive empirical study, we discover that the penalty of restricting decentralized spatial reasoning to natural language with limited budget is far smaller than suggested by prior two-agent literature. Nevertheless, we observe that substantial headroom still remains for multi-agent spatial reasoning in this setting, as evidenced by a significant gap between current centralized performance and an oracle best-of-$K$ ceiling. Moreover, we show that reinforcement learning with credit assignment substantially bolsters the decentralized model, effectively closing the gap with the centralized baseline.
\end{abstract}    
\begin{figure}[t]
  \centering
  \includegraphics[width=\columnwidth]{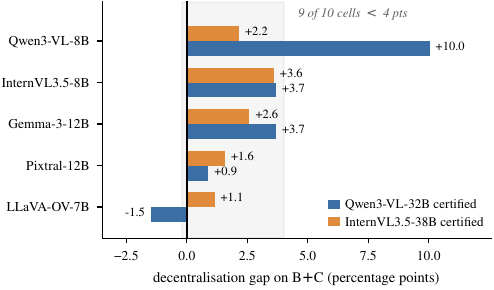}
  \caption{\textbf{The decentralization gap (centralized - decentralized) through language is narrow.} Accuracy lost against a centralised model that sees every view,
  on the synergistic slice (B+C) with a matched chain-of-thought listener. Nine of
  ten backbone/certifier cells fall below four points and LLaVA-OV-7B is negative. The
  one large gap is Qwen3-VL-8B scored on data certified by a model of its own family; it
  collapses from $+10.0$ to $+2.2$ points when only the certifier changes, while
  InternVL3.5-8B is unmoved.}
  \label{fig:teaser}
\end{figure}
\section{Introduction}
\label{sec:intro}

Spatial intelligence is fundamental to physical collaboration, enabling agents to synthesize disparate local observations into a cohesive shared mental model through natural language. However, existing multimodal spatial reasoning benchmarks overwhelmingly adopt a centralized paradigm, assuming an omniscient model with unconstrained visibility over all environment viewpoints~\cite{cho2026spatialclaw,gong2026s,yu2026evospatial,li2026spatialladder}. This assumption severely diverges from certain real-world distributed deployments such as multi-robot coordination, collaborative AR/VR, and interactive multi-agent systems~\cite{2900423.2900661,mandi2024roco,zhang2024building,chen2024scalable,park2023generative}. For instance, in V2V cooperative perception, single-agent 3D object detection suffers from occlusions and limited sensing ranges~\cite{Ma_2024_WACV} and streaming raw high-bandwidth visual observations across all nodes might be fundamentally unfeasible~\cite{mehr2026simbev2x}. Furthermore, in privacy-sensitive environments like home-assisted care, transferring raw camera feeds across personal domains is often restricted, whereas abstracted natural language exchange is permitted. Language thus serves as an essential scaffold in bandwidth and privacy-constrained settings, allowing agents to ground shared landmarks, infer relative orientations, and resolve spatial ambiguities across unobserved regions. As MLLMs evolve into active partners, there is an urgent need to systematically benchmark decentralized multi-view spatial reasoning across complex environments.
\\
\\
Given the scarcity of prior work in this regime, COSMIC~\cite{sikarwar2026communicating} stands as the closest antecedent to our setting and, to our
knowledge, the only systematic study of language-mediated spatial integration: two agents observe one indoor room from different egocentric viewpoints and exchange dialogue to solve a spatial query, with human-human dialogues collected for reference. While this serves as a valuable initial probe, it leaves the general problem untouched. 
We address this gap through the following main contributions\footnote{Codebase and Data splits will be made publicly available upon acceptance.}:

\begin{itemize}
  \itemsep0.2em
  \item \textbf{Task Generalization:} We generalize language-mediated spatial integration from a two-agent indoor room setting to an $N$-agent multi-view framework, introducing explicit per-message word budget and a zero-view listener to isolate communication quality from perception.
  \item \textbf{Dual-Certified Multi-View Benchmark:} We curate a benchmark of 10,007 QA pairs across six heterogeneous sources, introducing a dual-model certification protocol to strictly isolate genuinely synergistic multi-view spatial queries from text-only shortcuts.
  \item \textbf{A Holistic Exploration of Decentralization Cost:} Evaluating five backbones across 27 inference-only agentic design patterns reveals that the performance cost of language-mediated decentralization relative to a centralized pooled baseline is substantially narrower than prior two-agent literature implies.
  \item \textbf{High Oracle Ceiling \& RL Potential:} The best-of-$K$ oracle ceiling ($58.9\%$) exceeds the centralized pooled baseline ($38.7\%$) by over 20\%, demonstrating that natural language communication is not intrinsically lossy at this scale. Moreover, multi-agent RL with optimized credit assignment effectively strengthens decentralized communication to within 3.5\% of the centralized baseline.
\end{itemize}

Beyond these results, this paper contributes several methodological insights of lasting value which have been thoroughly discussed in Sec. \ref{sec:results}.
\section{Related Work}
\label{sec:related}

\paragraph{Multi-view spatial reasoning.}
A growing family of benchmarks probes whether multimodal models can reason over several
images of one scene rather than a single view, covering cross view correspondence,
egocentric-allocentric transfer, relative direction and distance, object permanence
across viewpoints, and embodied or robotic settings
\cite{yang2026mmsi,yeh2026seeing,zhang2026flatland,feng2025seeing,xu2026citycube,jeong2026learning,fu2024blink,yang2025thinking,wang2025spatial,ma20253dsrbench,tong2024cambrian,Chen_2024_CVPR,cheng2024spatialrgpt}. These studies have established that multi-view reasoning is substantially harder for current models than
single-image perception. They are, however, uniformly \emph{centralised}: one model is handed every view at once, so they measure perception and integration jointly and say nothing about what happens when the views are distributed across agents that cannot share pixels. We draw our question pool from six such sources  and then change the access structure rather than the
questions.
\paragraph{Cooperative perception with feature-level communication.}
In autonomous driving, cooperative perception has become a mature subfield: agents share
intermediate representations, usually in a common bird's-eye-view frame, to overcome
occlusion and limited range. Recent synthetic datasets coordinate up to eight connected
vehicles and four roadside units with multi-modal sensing and dense annotation, and the
accompanying architectures fuse agent features with attention
\cite{mehr2026simbev2x,zimmer2024tumtraf,li2022v2x,xu2022opv2v,yu2022dair,xu2022cobevt}. This literature is an existing demonstration that
multi-agent perception works, and it is the right comparison for scale. Its communication
channel, however, is a learned latent one, which requires that participants share an
architecture, a calibration and a clock. Our setting replaces that channel with natural
language, trading representational precision for interoperability, auditability and an
explicit bandwidth budget measured in words. Communication-efficient fusion imposes an analogous budget in the
feature channel\,\cite{hu2022where2comm}.

\begin{figure*}[t]
  \centering
  \includegraphics[width=\textwidth]{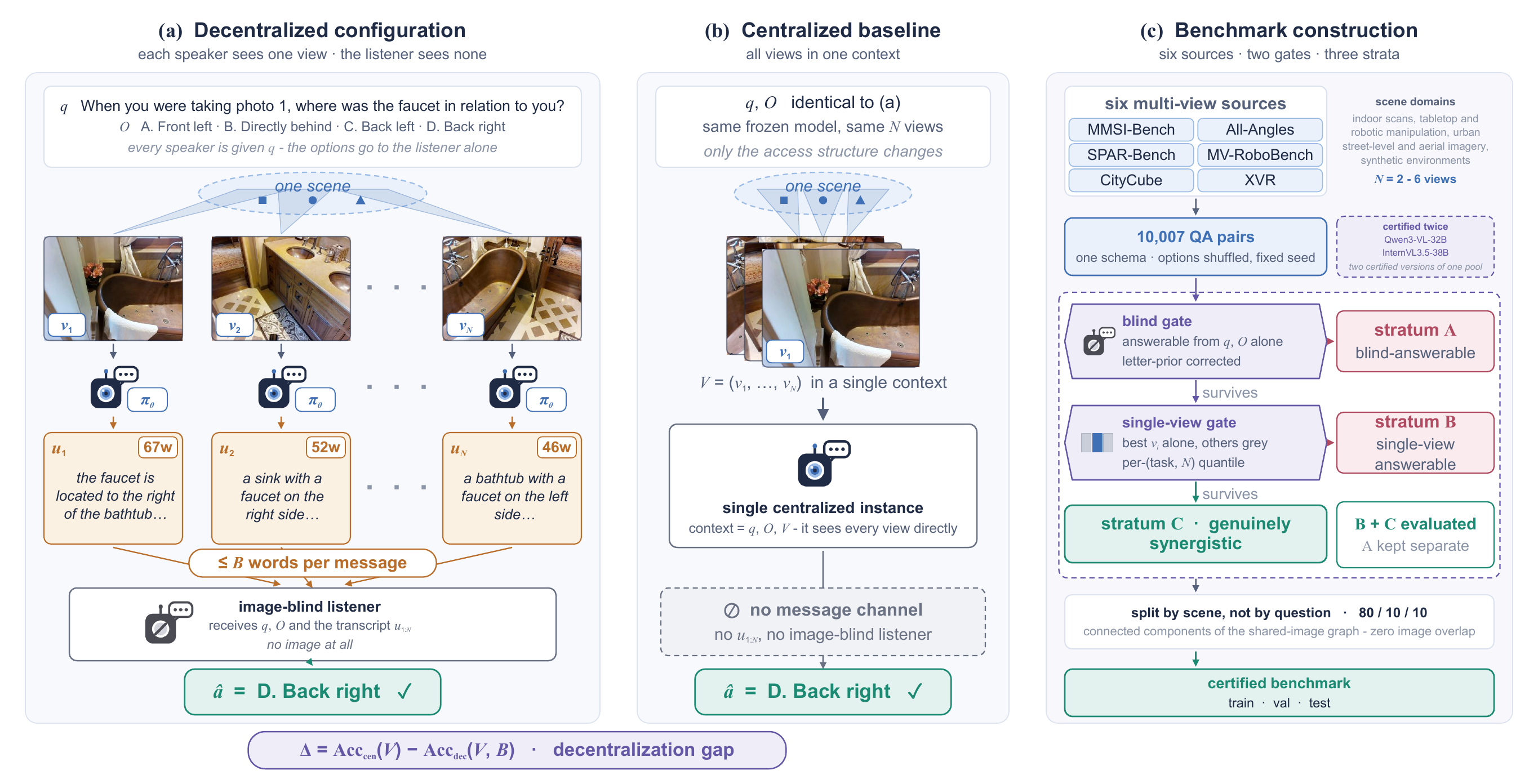}
  \caption{\textbf{(a) Decentralized}: $N$ speakers see one view $v_i$ each and the question
$q$, never the options, and send one message $u_i$ ($B \le 60$ words) to an image-blind
listener that answers from the transcript alone. \textbf{(b) Centralized}: the same frozen
model sees all $N$ views at once; the accuracy it gains is the decentralization gap
$\Delta$. \textbf{(c) Benchmark Construction}: six sources pass a blind and a single-view gate that
split the pool into strata A, B and C (counts in \cref{tab:composition}).}
  \label{fig:task}
\end{figure*}
\vspace{-1.5em}
\paragraph{Language-mediated collaboration between agents.}
Coordination in language rather than in features has a long history in
emergent-communication and referential-game research, where protocols are learned from
scratch and typically remain uninterpretable
\cite{lewis69,lazaridou2016multi,foerster2016learning,havrylov2017emergence,chaabouni2019anti,lowe2019pitfalls}. More recent work equips pretrained multimodal models with
dialogue and asks whether their existing linguistic competence transfers to collaborative
perception.\,\cite{du2023improving,li2023camel,wu2023autogen} While COSMIC~\cite{sikarwar2026communicating} established an initial benchmark for two-agent spatial dialogue in a single room, it represents the sole systematic investigation in this space. We depart from this constrained setup and generalize the problem along four key axes: First, $N$ is a variable
rather than fixed at two, which introduces the question of which partners to
reconcile with and makes per-agent credit assignment non-trivial. Second, our scenes span
six heterogeneous sources rather than indoor rooms alone. Third, messages are subject to
an explicit word budget, without which the cost of communication cannot be separated from
the cost of verbosity. Fourth, our listener sees no image, so that measured performance
reflects communication rather than a partially-informed answerer's own perception.

\vspace{-1.5em}

\paragraph{Credit assignment and post-training for multi-agent language policies.} Optimising several speakers against a single team reward is a credit-assignment problem of the kind long studied in cooperative multi-agent reinforcement learning, where
counterfactual and difference-reward estimators attribute outcomes to individual
contributors \cite{foerster2018counterfactual,rashid2020monotonic,sunehag2017value,lowe2017multi,yu2022surprising}. Policy-gradient post-training of language models has
converged on group-relative objectives that dispense with a learned critic
\cite{shao2024deepseekmath,liu2025understanding,Zheng2025GroupSP,Schulman2017ProximalPO,ahmadian-etal-2024-back,guo2025deepseek,liao2025improved}, and these have recently been extended to multi-turn and
multi-agent settings. Our contribution here is empirical rather than algorithmic: holding
the task, data, budget and evaluation fixed, we compare eight variants that differ in how
credit is distributed across speaker agents and in the importance-ratio granularity of the
update, and find that multi-agent reinforcement learning has substantial potential to
close or even surpass the decentralisation gap. The same comparison surfaces free-riding as a
concrete, measurable failure mode governed by the composition of the training pool.

\section{Decentralised Multi-View Spatial Reasoning}
\label{sec:task}

\subsection{Problem formulation}
\label{sec:task:formulation}

A scene comprises views $V=(v_1,\dots,v_N)$, question $q$, and options $\mathcal{O}$ ($|\mathcal{O}|=m$). We instantiate $N$ speaker agents and one listener. Speaker $i$ receives $q$, view $v_i$, and knowledge of $N-1$ hidden views (with no other views, peer outputs, or options), emitting a message $u_i$ (at most $B$ words). The listener receives $q$, $\mathcal{O}$, and transcript $u_{1:N}$ with \emph{no images}, outputting $\hat{a}\in\mathcal{O}$ (\Cref{fig:task}).

All $N{+}1$ roles share a frozen vision-language model, isolating routing effects from model capacity. The centralized baseline feeds a single model instance $q$, $\mathcal{O}$, and all $V$. We define the \textbf{decentralization gap} as
\begin{equation}
  \Delta = \mathrm{\textit{Acc}}_{\text{cen}}(V) - \mathrm{\textit{Acc}}_{\text{dec}}(V, B),
  \label{eq:gap}
\end{equation}
representing the accuracy penalty incurred by transmitting textual descriptions under a capacity constraint $B$ rather than directly feeding visual inputs.

Because messages are emitted simultaneously without peer access, speakers face a coordination problem under incomplete information. This frames the system as a decentralized partially observable decision process with an explicit communication channel, which governs credit assignment in \cref{sec:task:rl}.

\subsection{Multi-View Data Certification Protocol}\label{sec:task:certification}Evaluating the decentralization gap ($\Delta$) requires filtering out questions that do not genuinely demand multi-view integration. To isolate such cases, we partition the dataset into three strata, namely stratum A (blind-answerable), stratum B (single-view-answerable), and stratum C (genuinely synergistic), using a two-stage filtering pipeline. First, a blind gate removes questions answerable from language priors or option structure alone. To account for position bias over choice letters, a question is categorized as stratum A only if its imageless accuracy significantly exceeds the model's letter-prior expectation. Next, the single-view gate evaluates all surviving questions on each view in isolation, replacing all unobserved views with grey canvases of identical dimensions so that positional token geometry remains strictly preserved. To avoid the oracle bias of comparing a maximum across single views against a single centralized attempt, we apply a per-stratum quantile threshold within each (task, $N$) combination, marking a question as single-view-answerable (stratum B) if its top-performing isolated view clears this threshold. Finally, questions surviving both filtering stages constitute stratum C (genuinely synergistic). Because individual model certifiers possess inherent perceptual biases, we run this entire protocol independently twice using Qwen3-VL-32B~\cite{bai2025qwen3} and InternVL3.5-38B~\cite{wang2025internvl3}, reporting main evaluations across both certified sets to ensure our results are certifier-agnostic.

\subsection{Benchmark construction}
\label{sec:task:data}

\begin{table}[b]
  \centering\small
  \setlength{\tabcolsep}{4.2pt}
  \begin{tabular}{ccccccc}
    \toprule
    & & & \multicolumn{3}{c}{Stratum} & \\
    \cmidrule(lr){4-6}
    Certifier & Split & $n$ & A & B & C & B\,$+$\,C \\
    \midrule
    \multirow{3}{*}{Qwen3-VL-32B}
      & train & 8,040 & 2,499 & 1,658 & 3,883 & 5,541 \\
      & val   &   998 &   295 &   210 &   493 &   703 \\
      & test  &   969 &   292 &   205 &   472 &   677 \\
    \midrule
    \multirow{3}{*}{InternVL3.5-38B}
      & train & 8,015 & 2,444 & 1,728 & 3,843 & 5,571 \\
      & val   & 1,009 &   297 &   191 &   521 &   712 \\
      & test  &   983 &   288 &   174 &   521 &   695 \\
    \bottomrule
  \end{tabular}
  \caption{\textbf{Certified splits.} The same 10,007 questions partitioned twice, by two
  certifiers of different families. Splits (80/10/10) are taken over \emph{scenes}, giving zero image
  overlap between train, val and test. }
  \label{tab:composition}
\end{table}

\begin{figure}[h]
  \centering
  \includegraphics[width=\columnwidth]{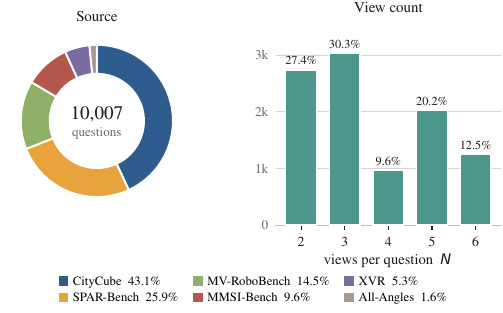}
  \caption{\textbf{Pool composition.} Six multi-view sources contribute 10,007 questions
  spanning $N{=}2$ to $6$ views.}
  \label{fig:composition}
  \vspace{-2em}
\end{figure}

No single benchmark supplies enough genuinely multi-view questions across a range of $N$,
so we amalgamate six sources, MMSI-Bench~\cite{yang2026mmsi}, All-Angles
Bench~\cite{yeh2026seeing}, SPAR-Bench~\cite{zhang2026flatland}, MV-RoboBench~\cite{feng2025seeing},
CityCube~\cite{xu2026citycube} and XVR~\cite{jeong2026learning}, spanning indoor scans, tabletop and robotic
manipulation, urban street-level and aerial imagery, and synthetic environments with $N$
from $2$ to $6$ (\cref{fig:composition}). We normalise every source to a common schema of
ordered view list, question, option set and answer, and discard questions whose views
cannot be recovered as separate images, giving $10{,}007$ question and answer pairs whose
composition is in \cref{tab:composition}. Because the blind gate measures exploitation of
option structure, we also shuffle the option order of every question with a fixed seed and
record the permutation, so the letter-prior correction of \cref{sec:task:certification} is
applied against a pool with no positional signal to exploit. The pipeline is visually represented in Fig. \ref{fig:task}.

\paragraph{Splitting by scene, not by question.}
Several sources contribute many questions about the same scene, so a question-level split
would place different questions about identical images on both sides of the train/test
boundary. We build a graph in which two questions are joined when they share any image
file and take connected components as scenes, then assign whole components to splits (80/10/10). This yields $8{,}040$ training, $998$ validation and $969$ test questions with zero
image overlap between them, verified exhaustively. Stratum proportions in each split match
the pool to within one point. Because the two certifiers induce different A/B/C labels,
their splits differ; all training pools used in \cref{sec:task:rl} are additionally
filtered against the union of both test sets' scenes, so that a single trained policy can
be evaluated on either certified set without leakage.

\subsection{Inference-only agentic designs}
\label{sec:task:designs}
Before training anything, we ask how far the gap closes by prompting and protocol alone. Constrained by finite compute resources, we restrict our study to $27$ representative inference-only designs, all respecting the constraints of
\cref{sec:task:formulation}, so that an agent sees only its own view, communicates only in
natural language, and is held to a word budget. They fall into eight families.
\textbf{Decision rule} varies what a speaker produces, either an answer plus a
justification for a view-free judge (\textsc{justify}) or the option to decline when
uncertain (\textsc{abstain}) \cite{feng2024don}. \textbf{Topology} varies who hears whom, from a round-robin
\textsc{chain} through a \textsc{ring} and a $3/5$-\textsc{regular} graph to full broadcast
(\textsc{dense}) and a two-level cluster \textsc{hier}archy \cite{li2024improving}. \textbf{Rounds} varies how long
exchange continues, over two rounds (\textsc{dense}), four (\textsc{dense4}) or an
\textsc{adaptive} schedule halting on consensus \cite{chen2024reconcile}. \textbf{Initiative} lets agents direct the
exchange through a single question (\textsc{query}), a question-answer-followup triple
(\textsc{ask2}), or mutual interrogation (\textsc{p2p}) \cite{das2017learning}. \textbf{Content} constrains what a
message may contain, by a fixed \textsc{schema}, assigned \textsc{roles} over objects,
geometry and counting, \textsc{slots} partitioned across speakers, or
claim-confidence-evidence triples (\textsc{cce}) \cite{hong2024metagpt}. \textbf{Bandwidth} ablation sweeps the budget at
$16$, $32$ and $128$ words, a fixed \textsc{total\_budget} divided $N$ ways, and an adaptive
\textsc{bid} in which agents compete for words. \textbf{Tools} grant a capability beyond
description, an own-view crop (\textsc{zoom}) and a structured object list with positions
(\textsc{ground}) \cite{wu2024v}. A final group varies \textbf{aggregation} on the listener side rather
than the speakers, through chain-of-thought, self-consistency over samples, truncation to
$32$ words, and a heterogeneous ensemble. We deliberately exclude designs that would share
camera pose or a common coordinate frame, since those leak the geometry the task is meant
to test.

\subsection{Learning to communicate}
\label{sec:task:rl}

The decentralised system is differentiable only through the speakers, and the only
supervision available is whether the listener answers correctly. This is a cooperative
multi-agent policy-gradient problem with a shared, sparse, terminal reward. We fine-tune a
LoRA adapter~\cite{hu2021lora} shared across all $N$ speaker roles, keep the listener frozen, and
use group-relative objectives that dispense with a learned critic~\cite{shao2024deepseekmath}.
\paragraph{Credit assignment.}
For a question we sample $K$ message sets
$u^{(k)}_{1:N}\sim\pi_\theta(\cdot\mid q,v_{1:N})$ and score each with the frozen scoring listener, which assigns probability to the option letters in one forward pass. Listener success is the standard reward in referential games~\cite{lazaridou2016multi,havrylov2017emergence}; we take its continuous value on the gold option rather than a binary outcome, for a denser credit signal~\cite{lightman2024let},
\begin{equation}
  R_k \;=\; p_{\text{lis}}\!\left(a^\star \mid q,\mathcal{O},u^{(k)}_{1:N}\right),
  \label{eq:reward}
\end{equation}
and calculate the team advantage via group relative normalization following GRPO~\cite{shao2024deepseekmath}:
\begin{equation}
  A^{\text{team}}_k \;=\; \frac{R_k-\mu_R}{\sigma_R+\epsilon},\qquad
  \mu_R=\tfrac{1}{K}\textstyle\sum_j R_j,
  \label{eq:adv}
\end{equation}
with $\sigma_R$ denoting the within-group standard deviation; the Dr.~GRPO variant~\cite{liu2025understanding}
omits it, removing the question-level difficulty bias it induces, and also drops the
token-count normalisation responsible for the length bias.

Equation~\eqref{eq:adv} gives every speaker the same scalar, so a message that contributed
nothing is reinforced exactly as much as one that carried the answer. Let $R^{-i}_k$ be the reward obtained when speaker $i$'s message is withheld
and the others are kept, and $Q^{i,k'}_k$ for the reward when it is replaced by the message
the same speaker produced in a different sampled set $k'$. We compare three
estimators. Joint credit and removal credit share an additive form,
\begin{align}
  A^{\text{joint}}_{i,k} &= A^{\text{team}}_k, \label{eq:joint}\\
  A^{\text{rem}}_{i,k}   &= A^{\text{team}}_k \;+\; \lambda\,
     \frac{R_k - R^{-i}_k}{\sigma_{c}+\epsilon}, \label{eq:loo}
\end{align}
in which $\lambda$ weights a per-speaker correction scaled by $\sigma_{c}$, the standard
deviation of the removal deltas within the group, and $\lambda{=}0$ recovers joint credit
exactly, so the two are one estimator at two settings. The term $R_k-R^{-i}_k$ is the
difference reward of cooperative MARL~\cite{foerster2018counterfactual}. Substitution credit instead replaces the team
term with a baseline of RLOO form~\cite{ahmadian-etal-2024-back} taken not over independent
rollouts but over rollouts in which only that speaker's message is substituted,
following C3~\cite{chen2026exact},
\begin{equation}
  A^{\text{sub}}_{i,k} \;=\; R_k \;-\; \frac{1}{K-1}\sum_{k'\neq k} Q^{i,k'}_k .
  \label{eq:sub}
\end{equation}
Equation~\eqref{eq:loo} measures a speaker against its own silence, which changes how many
observers the listener believes it has, whereas \eqref{eq:sub} measures it against an
alternative utterance from the same speaker, holding the observer count fixed.
\cref{sec:results:rl} shows this choice matters more than any other we make. Removal credit
also underlies CCPO~\cite{li2026counterfactual}, which calibrates advantages against global
rollout history; we use the within-group form.
\paragraph{Objective.}
Let $\rho_{i,k,t}(\theta)=\pi_\theta(u_{i,k,t}\mid\cdot)/\pi_{\theta_{\text{old}}}(u_{i,k,t}\mid\cdot)$
be the per-token importance ratio over the $T_{i,k}$ tokens of message $u_{i,k}$. Token-level
methods clip each $\rho_{i,k,t}$ independently, whereas GSPO~\cite{Zheng2025GroupSP} forms a single
length-normalised sequence ratio
\begin{equation}
  s_{i,k}(\theta)\;=\;\Big(\textstyle\prod_{t=1}^{T_{i,k}}\rho_{i,k,t}(\theta)\Big)^{1/T_{i,k}}
\;=\;\exp\!\big(\ell^{\text{old}}_{i,k}-\ell_{i,k}\big),
  \label{eq:seqratio}
\end{equation}
with $\ell$ the mean token negative log-likelihood, and clips that. Writing $c(\cdot)$ for
the clipping operator~\cite{Schulman2017ProximalPO}, applied to the sequence ratio under
GSPO's asymmetric bounds $[1-\varepsilon_{\text{lo}}, 1+\varepsilon_{\text{hi}}]$ and to each
$\rho_{i,k,t}$ otherwise, the objective is
\begin{equation}
  \mathcal{L}(\theta)= -\,\mathbb{E}\Big[\min\big(s_{i,k}A_{i,k},\,
    c(s_{i,k})A_{i,k}\big)\Big] \;+\; \beta\,\mathrm{KL}\!\left[\pi_\theta\,\|\,\pi_{\text{ref}}\right].
  \label{eq:obj}
\end{equation}
Our eight variants cross the credit families of
\eqref{eq:joint} to \eqref{eq:sub} with the ratio granularity and normalisation of
\eqref{eq:adv} and \eqref{eq:seqratio}, including a hybrid applying removal credit under a
sequence-level ratio. Task, data, budget and evaluation are fixed across all eight, so each
comparison isolates one choice.
\paragraph{Training pool composition.}
Since stratum A questions are answerable without any view, including them rewards a policy
for outcomes its messages did not cause. We therefore train on B+C only. \cref{sec:supp:freeride} reports the controlled comparison against A+B+C, which is the
clearest available probe of free-riding.


\section{Evaluation and Results}
\label{sec:results}

\begin{table*}[t]
  \centering\footnotesize
  \setlength{\tabcolsep}{5pt}
  \renewcommand{\arraystretch}{0.95}
  \begin{tabular}{lllcccc}
    \toprule
    & & & \multicolumn{2}{c}{\textbf{Chain-of-Thought listener}}
        & \multicolumn{2}{c}{\textbf{Scoring listener}} \\
    \cmidrule(lr){4-5}\cmidrule(lr){6-7}
    Backbone & Certifier & System & ALL & B\,$+$\,C & ALL & B\,$+$\,C \\
    \midrule
    \multirow{4}{*}{Qwen3-VL-8B} & \multirow{2}{*}{Qwen3-VL-32B} & centralised & 42.6 & 38.7 & 39.4 & 35.2 \\
    & & decentralised & 35.4\,{\scriptsize$(-7.2)$} & \cellcolor{softblue}{\textbf{28.7\,{\scriptsize$(-10.0)$}}} & 36.2\,{\scriptsize$(-3.2)$} & \cellcolor{softblue}{\textbf{29.1\,{\scriptsize$(-6.1)$}}} \\
    \cmidrule(l){2-7}
     & \multirow{2}{*}{InternVL3.5-38B} & centralised & 38.8 & 33.7 & 37.3 & 31.5 \\
    & & decentralised & 35.6\,{\scriptsize$(-3.2)$} & 31.5\,{\scriptsize$(-2.2)$} & 34.2\,{\scriptsize$(-3.1)$} & 30.1\,{\scriptsize$(-1.4)$} \\
    \midrule
    \multirow{4}{*}{InternVL3.5-8B} & \multirow{2}{*}{Qwen3-VL-32B} & centralised & 41.0 & 35.9 & 40.5 & 37.1 \\
    & & decentralised & 38.0\,{\scriptsize$(-3.0)$} & 32.2\,{\scriptsize$(-3.7)$} & 37.8\,{\scriptsize$(-2.7)$} & 31.3\,{\scriptsize$(-5.8)$} \\
    \cmidrule(l){2-7}
     & \multirow{2}{*}{InternVL3.5-38B} & centralised & 37.1 & 31.9 & 37.2 & 31.2 \\
    & & decentralised & 34.0\,{\scriptsize$(-3.1)$} & 28.3\,{\scriptsize$(-3.6)$} & 34.9\,{\scriptsize$(-2.3)$} & 28.8\,{\scriptsize$(-2.4)$} \\
    \midrule
    \multirow{4}{*}{LLaVA-OV-7B} & \multirow{2}{*}{Qwen3-VL-32B} & centralised & 35.6 & 30.7 & 38.1 & 33.1 \\
    & & decentralised & 37.0\,{\scriptsize$(+1.4)$} & \cellcolor{softred}{\textbf{32.2\,{\scriptsize$(+1.5)$}}} & 36.5\,{\scriptsize$(-1.6)$} & 31.9\,{\scriptsize$(-1.2)$} \\
    \cmidrule(l){2-7}
     & \multirow{2}{*}{InternVL3.5-38B} & centralised & 34.8 & 31.1 & 36.4 & 31.1 \\
    & & decentralised & 34.0\,{\scriptsize$(-0.8)$} & 29.9\,{\scriptsize$(-1.2)$} & 36.2\,{\scriptsize$(-0.2)$} & \cellcolor{softred}{\textbf{30.9\,{\scriptsize$(-0.2)$}}} \\
    \midrule
    \multirow{4}{*}{Pixtral-12B} & \multirow{2}{*}{Qwen3-VL-32B} & centralised & 35.3 & 30.3 & 38.1 & 33.5 \\
    & & decentralised & 34.8\,{\scriptsize$(-0.5)$} & 29.4\,{\scriptsize$(-0.9)$} & 36.2\,{\scriptsize$(-1.9)$} & 29.4\,{\scriptsize$(-4.1)$} \\
    \cmidrule(l){2-7}
     & \multirow{2}{*}{InternVL3.5-38B} & centralised & 33.9 & 29.1 & 37.6 & 29.4 \\
    & & decentralised & 31.1\,{\scriptsize$(-2.8)$} & 27.5\,{\scriptsize$(-1.6)$} & 31.6\,{\scriptsize$(-6.0)$} & 26.6\,{\scriptsize$(-2.8)$} \\
    \midrule
    \multirow{4}{*}{Gemma-3-12B} & \multirow{2}{*}{Qwen3-VL-32B} & centralised & 38.8 & 34.9 & 39.8 & 36.2 \\
    & & decentralised & 35.9\,{\scriptsize$(-2.9)$} & 31.2\,{\scriptsize$(-3.7)$} & 36.8\,{\scriptsize$(-3.0)$} & 33.1\,{\scriptsize$(-3.1)$} \\
    \cmidrule(l){2-7}
     & \multirow{2}{*}{InternVL3.5-38B} & centralised & 35.7 & 30.8 & 37.0 & 31.5 \\
    & & decentralised & 31.5\,{\scriptsize$(-4.2)$} & 28.2\,{\scriptsize$(-2.6)$} & 32.5\,{\scriptsize$(-4.5)$} & 29.2\,{\scriptsize$(-2.3)$} \\
    \bottomrule
  \end{tabular}
  \caption{\textbf{Accuracy (\%) across 5 backbones.} The parentheses give the
  decentralised change against its own centralised counterpart. In the B\,$+$\,C columns, 
  {\color{softred}\rule{2mm}{2mm}} denotes the smallest negative gaps (or positive ones), and 
  {\color{softblue}\rule{2mm}{2mm}} denotes largest negative gaps.}
  \label{tab:gapmatrix}
\end{table*}

\subsection{Experimental setting}
\label{sec:results:setup}
We evaluate five open-weight vision-language models spanning three families and two
parameter scales, Qwen3-VL-8B~\cite{bai2025qwen3}, InternVL3.5-8B~\cite{wang2025internvl3},
LLaVA-OneVision-7B~\cite{li2024llava}, Pixtral-12B~\cite{agrawal2024pixtral} and
Gemma-3-12B~\cite{team2025gemma}. Within a configuration one model plays speaker, listener and centralised reference, so
every number is an internal comparison, and the two certifiers, Qwen3-VL-32B and
InternVL3.5-38B, are larger models of different families so no evaluated backbone certifies
itself. Unless stated otherwise, $B{=}60$ words serves as a soft constraint, speakers decode greedily, and the listener
runs either as a \emph{chain-of-thought} listener reasoning in at most $512$ tokens or a
\emph{scoring} listener returning the highest-probability option. Both arms always use the
same listener, giving CoT-matched and score-matched gaps. Hidden views are replaced by grey
canvases of identical size rather than removed, so image-token count is constant and no gap
can arise from differing context length (\cref{sec:supp:setup}).

\subsection{Computational resources}
\label{sec:results:compute}
All experiments ran on an NVIDIA GH200 cluster using $8$ nodes per inference or
reinforcement-learning job. Cost by experiment
category is reported in \cref{sec:supp:compute} to support replication at known expense. 

\subsection{The cost of decentralisation}
\label{sec:results:gap}
Three reference points frame every number that follows. The \textbf{centralised} bar is the
same backbone given all views, the \textbf{untrained} bar is the task as posed with no
training, and the \textbf{oracle best-of-$K$} ceiling keeps, per question, the sampled
message set the listener scores best.
\cref{tab:gapmatrix} reports the central measurement of the paper.
\textbf{Routing cross-view evidence through language costs far less than the
two-agent literature implies}, with nine of the ten CoT-matched B+C changes within four
points of zero. The single large loss of $-10.0$ points belongs to Qwen3-VL-8B on the
Qwen-certified set, and \mbox{LLaVA-OneVision-7B} is the one positive cell at $+1.5$
points, its decentralised system outperforming the same weights given every view.
The matrix also shows the gap depends on who certified the data. Qwen3-VL-8B's B+C loss
shrinks from $-10.0$ to $-2.2$ when only the certifier changes, while
InternVL3.5-8B's is unmoved at $-3.7$ to $-3.6$, each family showing its largest gap on
data certified by its own (\cref{fig:teaser}). A certifier assigns to stratum C what
\emph{it} cannot answer, selecting a harder slice for models resembling it, and our two
agree on only $60.9\%$ of questions despite near-identical marginals. \textbf{A gap
reported against a single model-certified benchmark is partly a property of that
certifier}, so the fix is to certify twice with unrelated family backbones.

\subsection{Headroom, and what closes it}
\label{sec:results:rl}
\label{sec:results:ceiling}
A narrow gap admits two explanations, that the channel is nearly lossless or that our
designs fail to exploit it. The oracle best-of-$K$ ceiling separates them, reaching $58.9$
on B+C and $63.9$ on ALL against the centralised $38.7$ and $42.6$ (\cref{tab:rl}) respectively.
Selection uses the gold answer, so this is an upper bound, but it shows the information
reaches language far more often than any single rollout recovers.
\textbf{Language at $B$ words is not the bottleneck; eliciting the right $B$ words is.}
The ceiling is $75.3$ on stratum A and $75.1$ on stratum B but only $51.9$ on stratum C, so upper bound is smallest
where the task is hardest (\cref{sec:supp:ceiling}), making this a learning problem.

\begin{table}[t]
  \centering\footnotesize
  \setlength{\tabcolsep}{4pt}
  \renewcommand{\arraystretch}{0.95}
  \begin{tabular}{@{}cl ccccc@{}}
    \toprule
    & & ALL & A & B & C & B\,$+$\,C \\
    \midrule
    \multirow{3}{*}{\rotatebox[origin=c]{90}{\footnotesize\textsc{bars}}}
      & centralised   & 42.6 & 51.7 & 47.8 & 34.8 & 38.7 \\
      & untrained   & 35.4 & 51.0 & 35.1 & 25.9 & 28.7 \\
      & oracle best-of-$K$        & 63.9 & 75.3 & 75.1 & 51.9 & 58.9 \\
    \midrule
    \multirow{3}{*}{\rotatebox[origin=c]{90}{\footnotesize\textsc{joint}}}
      & GRPO       & 38.7 & 48.0 & 44.9 & 30.3 & 34.7 \\
      & GSPO     & 35.8 & 44.2 & 42.0 & 28.0 & 32.2 \\
      & Dr.\ GRPO& 34.3 & 49.7 & 36.6 & 23.7 & 27.6 \\
    \midrule
    \multirow{3}{*}{\rotatebox[origin=c]{90}{\footnotesize\textsc{rem}}}
      & Leave-one-out (LOO)  & 38.9 & 47.6 & 45.9 & 30.5 & \cellcolor{hirow}\best{35.2} \\
      & GSPO\,$+$\,LOO            & 38.6 & 49.3 & 44.4 & 29.5 & 34.0 \\
      & Dr.\ GRPO\,$+$\,CF & 38.5 & 49.3 & 39.5 & 31.4 & 33.8 \\
    \midrule
    \multirow{2}{*}{\rotatebox[origin=c]{90}{\footnotesize\textsc{sub}}}
      & C3 substitution~\cite{chen2026exact} & 29.5 & 47.3 & 26.3 & 19.9 & 21.9 \\
      & Leave-one-out \textit{(pool ABC)} & 35.6 & 55.1 & 34.2 & 24.2 & 27.2 \\
    \bottomrule
  \end{tabular}

  \caption{\textbf{Credit assignment variants.}
  Accuracy (\%), final checkpoint, CoT listener; best B\,$+$\,C cell shaded. Last row
  widens the pool to A+B+C (\cref{sec:supp:freeride}).}
  \label{tab:rl}
\end{table}

\begin{table}[b]
  \centering\small
  \setlength{\tabcolsep}{6pt}
  \renewcommand{\arraystretch}{1.0}
  \begin{tabular}{cccc}
    \toprule
    & & \multicolumn{2}{c}{\textbf{Evaluated on}} \\
    \cmidrule(lr){3-4}
    \textbf{Method} & \textbf{Train Pool} & \textbf{Qwen-cert.} & \textbf{IVL-cert.} \\
    \midrule
    \emph{Centralised} & -- & 37.1 & 31.2 \\
    \emph{Untrained}   & -- & 31.3 & 28.8 \\
    \midrule
    \multirow{2}{*}{\textbf{Leave-one-out (LOO)}} 
      & Qwen-cert. & 32.9 & \cellcolor{hirow}\textbf{30.9} \\
      & IVL-cert.  & 31.8 & 28.3 \\
    \midrule
    \multirow{2}{*}{\textbf{GSPO}} 
      & Qwen-cert. & 34.3 & 30.4 \\
      & IVL-cert.  & \cellcolor{hirow}\textbf{35.6} & 30.5 \\
    \bottomrule
  \end{tabular}
  \caption{\textbf{Transfer across backbone and certifier.} B\,$+$\,C accuracy (\%),
  InternVL3.5-8B, scoring listener. Columns give the evaluation set's certifier, rows the
  training pool's; best trained cell per column shaded. CoT cells in \cref{sec:supp:transfer}.}
  \label{tab:transfer}
\end{table}

\cref{tab:rl} compares the eight variants of \cref{sec:task:rl} under one protocol. The
best two variants, removal (leave-one-out) at $35.2$ and joint at $34.7$, lie within
$0.5$ points of each other on B+C against the centralized $38.7$ and untrained $28.7$,
closing $60$-$65\%$ of the gap from only $5{,}541$ questions. Substitution credit reaches only $21.9$, which is below the untrained baseline, so scoring a speaker against
a counterfactual utterance is actively harmful. Changing the optimiser while holding credit
fixed moves far less, with sequence-level ratios at $32.2$.
Two caveats belong here. The strongest variant emits $108$ words against a budget of $60$
and the trainer applies no length penalty, so part of its advantage is bought with
bandwidth the task does not grant. Among budget-respecting variants joint credit leads at
$34.7$ with $48.6$ words (\cref{sec:supp:rl}).
\cref{tab:transfer} asks whether this is a property of one backbone. We repeat one
variant from each family on InternVL3.5-8B, training on each certified pool and scoring on both
test sets, each pool filtered against the union of both sets' scenes to prevent leakage,
and report the scoring listener because it defines the reward of \cref{eq:reward}. Three
results follow. The signal transfers with trained policies exceeding the untrained bar on a
backbone that the recipe was never tuned for, with the best cell closing most of the centralized
gap. The ranking, however, is not preserved across backbones, and this experiment is
powered to test transfer rather than to rank schemes. We therefore read \cref{tab:rl} as
backbone-specific and not a portable ranking; per-cell
numbers and the full variant set are in \cref{sec:supp:transfer}.

\vspace{-0.1em}
\subsection{What prompting and protocol alone can buy}
\label{sec:results:designs}
\vspace{-0.6em}
\begin{table}[h]
  \centering\footnotesize
  \setlength{\tabcolsep}{3.2pt}
  \renewcommand{\arraystretch}{0.95}
  \begin{tabular}{@{}cl ccccc@{}}
    \toprule
    & & ALL & A & B & C & B\,$+$\,C \\
    \midrule
    \multirow{4}{*}{\rotatebox[origin=c]{90}{\footnotesize\textsc{bars}}}
      & centralised (CoT)         & 42.6 & 51.7 & 47.8 & 34.8 & 38.7 \\
      & centralised (score)       & 39.4 & 49.3 & 51.7 & 28.0 & 35.2 \\
      & untrained (score)         & 36.2 & 52.7 & 42.0 & 23.5 & 29.1 \\
      & untrained (CoT)           & 35.4 & 51.0 & 35.1 & 25.9 & 28.7 \\
    \midrule
    \multirow{5}{*}{\rotatebox[origin=c]{90}{\footnotesize\textsc{topology}}}
      & \textsc{chain}            & 38.0 & 55.5 & 40.0 & 26.3 & 30.4 \\
      & \textsc{ring}             & 37.7 & 54.8 & 46.8 & 23.1 & 30.3 \\
      & \textsc{regular}          & 36.1 & 51.0 & 46.8 & 22.3 & 29.7 \\
      & \textsc{dense}            & 36.5 & 54.5 & 46.3 & 21.2 & 28.8 \\
      & \textsc{hier}             & 35.5 & 53.4 & 42.0 & 21.6 & 27.8 \\
    \midrule
    \multirow{6}{*}{\rotatebox[origin=c]{90}{\footnotesize\textsc{others}}}
      & \textsc{dense4} \textit{(rounds)}   & 35.8 & 53.4 & 42.4 & 22.0 & 28.2 \\
      & \textsc{p2p} \textit{(initiative)} & 38.6 & 55.8 & 45.9 & 24.8 & 
      \cellcolor{hirow}\best{31.2} \\
      & \textsc{words128}$^\dagger$ & 38.8 & 54.8 & 48.8 & 24.6 & 31.9 \\
      & \textsc{agg\_cot} \textit{(aggregation)} & 38.0 & 54.1 & 39.0 & 27.5 & 31.0 \\
      & \textsc{abstain} \textit{(decision)} & 27.9 & 58.9 & 21.5 & 11.4 & 14.5 \\
    \bottomrule
  \end{tabular}

  \caption{\textbf{Inference-only designs} Accuracy (\%), $969$ test
  questions; all $27$ in \cref{sec:supp:designs}. Best admissible
  B\,$+$\,C cell shaded. $^\dagger$ denotes budget exceeded. }
  \label{tab:designs}
\end{table}
\begin{figure*}[t]
  \centering
  \includegraphics[width=\textwidth]{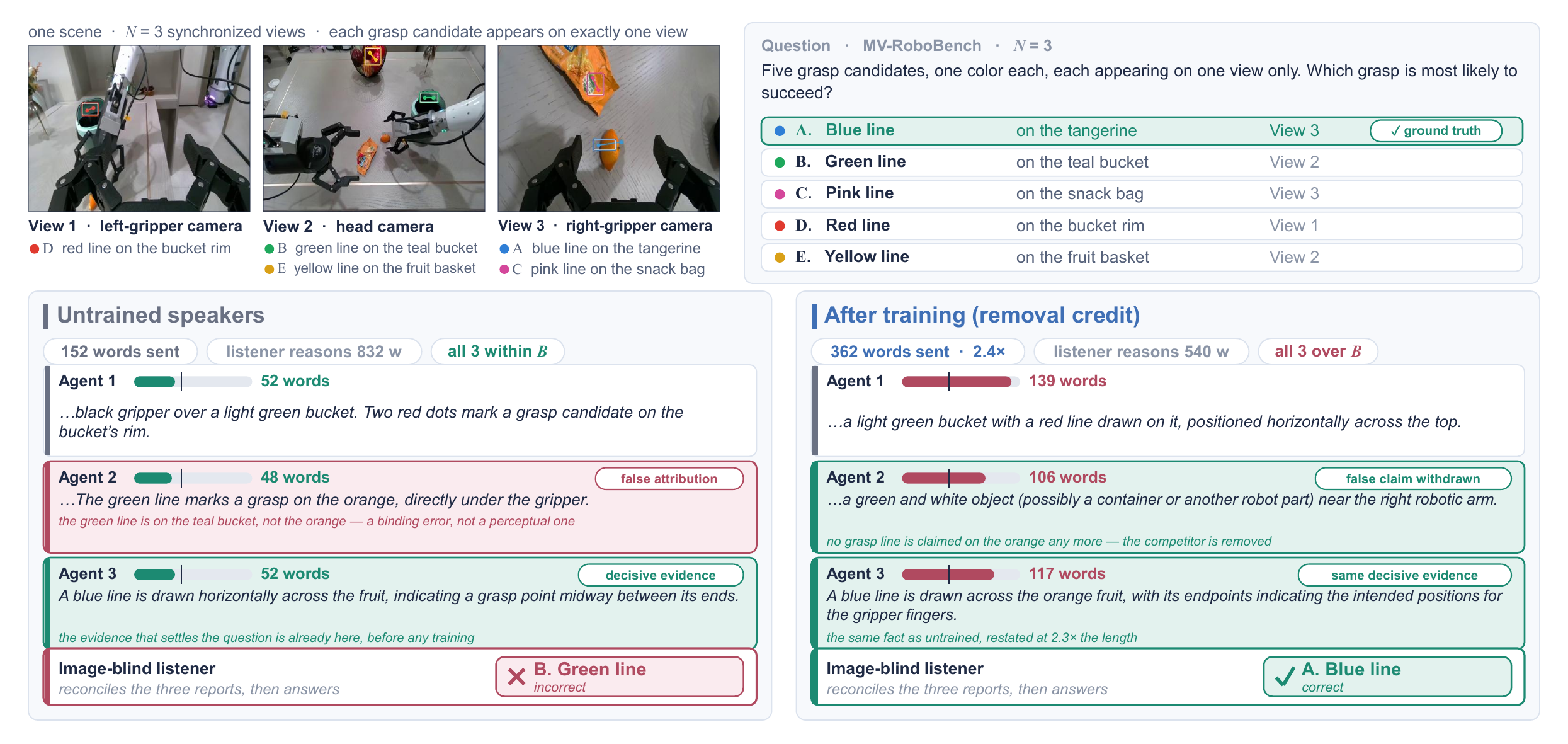}
  \caption{\textbf{Training resolves a false attribution, but overspends the budget.} Removal credit corrects the listener from B to A by eliminating a false competitor rather than adding new evidence (increasing word count from $152 \rightarrow 362$ words). Grasp-candidate bounding boxes are visualized in the source views solely for illustrative purposes and are not provided as inputs to the model.}
  \label{fig:qual}
\end{figure*}

\cref{tab:designs} reports the designs of \cref{sec:task:designs} that carry our
observations. Across all $27$, seventeen improve on the untrained bar for their listener
and ten fall below it, but the best admissible design reaches only $31.2$ on B+C against a
centralised $38.7$, recovering roughly a fifth of the gap. Prompting and protocol alone
therefore do not close it. \textbf{Protocol design recovers about a fifth of the gap; the
rest is a learning problem.}
The structure of the failures is more informative than the magnitudes. Topology is monotone
in sparsity, from \textsc{chain} at $30.4$ through \textsc{ring} $30.3$,
$3/5$-\textsc{regular} $29.7$ and \textsc{dense} $28.8$ to \textsc{hier} $27.8$, the last
two below speaking once, and four rounds adds nothing at $28.2$. We read this as crowding,
since every message a speaker receives describes a view it cannot verify and dilutes the
one it can report reliably. The exception is \textsc{p2p} at $31.2$, where agents
interrogate one another, suggesting a second round usefully carries a targeted question
rather than more description. \textsc{abstain} at $14.5$ reflects that declining
scores as wrong. While higher bandwidth improves performance, task constraints and finite compute budgets make multi-round and dependent protocols (such as \textsc{p2p}) impractical to scale during training. To maximize evaluation throughput under restricted compute, we select a single-round configuration with a scoring listener, which minimizes sampling costs while maintaining a clean training substrate.

\subsection{Qualitative analysis}
\label{sec:results:qual}

As demonstrated in \cref{fig:qual}, untrained speakers caption their view without reference to the question and state spatial
attributions with a confidence it does not support, whereas trained speakers lead with the
question-relevant relation and name anchors likely visible to others. The failure isolated
there is not perceptual: agent~2 sees a green line and sees an orange, and the error lies
in binding one to the other. A sighted listener would reject the claim, a blind one cannot,
and the attribution propagates into the answer. The dual failure, correspondence, also appears: two speakers describe one object as two, and the listener double-counts it (\cref{sec:supp:qual}).

\section{Conclusion}
\label{sec:conclusion}



In this work, we present a rigorous benchmark framework to quantify the cost of replacing shared visual representations with budget-constrained natural language. We introduce a novel spatial reasoning paradigm in which N agents each hold distinct viewpoints, communicate under strict word limits, and must guide an image-blind listener to a correct spatial decision. To enable honest measurement, we curate and double-certify questions across blind-answerable, single-view, and synergistic strata
using independent certifying architectures.
Beyond dataset formulation and stratified certification, we evaluate communication limits and show that existing models forfeit far less performance than previously assumed, with the strongest backbone matching its centralized counterpart and most pairs within four points of it. Finally, we expose the underlying failure modes of multi-agent communication and introduce
an attribution-based training paradigm with a shared reward structure that recovers roughly two-thirds of the remaining performance gap. Through our stratified dataset, evaluation protocol, and training insights (Sec. \ref{sec:results:gap} - \ref{sec:results:qual}), we set a new standard for language-mediated multi-view spatial reasoning.

\section{Limitations}
\label{sec:limitations}

Three limitations remain. First, synergy is relative to the chosen certifiers. A question is synergistic because two specific large models failed to answer it blind or from one view, yet another family might disagree and a stronger closed-source model might solve unassisted what ours could not. Two unrelated certifiers bound this. Second, we do not enforce a hard word budget during training (a soft constraint): the reward carries no length penalty. A hard cap or a length-penalized reward is the natural next experiment. Third, all agents share one backbone. In most experiments, every speaker and listener is an instance of the same frozen model, which isolates the access structure but sets aside heterogeneous teams.

\section{LLM Usage Statement}
\label{sec:llm}

An LLM was used to polish the writing of this manuscript, including grammar,
phrasing, and consistency of terminology.

{
    \small
    \bibliographystyle{ieeenat_fullname}
    \bibliography{main}

@String(CVPR= {IEEE Conf. Comput. Vis. Pattern Recog.})

@String(ICCV= {Int. Conf. Comput. Vis.})

@String(AAAI = {AAAI})

@String(CVPR  = {CVPR})

@String(ICCV  = {ICCV})

@article{sikarwar2026communicating,
  title={Communicating about Space: Language-Mediated Spatial Integration Across Partial Views},
  author={Sikarwar, Ankur and Mishra, Debangan and Nikhil, Sudarshan and Kumaraguru, Ponnurangam and Agrawal, Aishwarya},
  journal={arXiv preprint arXiv:2603.27183},
  year={2026}
}

@article{mehr2026simbev2x,
  title={SimBEV2X: A Large-Scale Dataset and Data Generation Tool for Multi-Task Vehicle-to-Everything Cooperative Perception},
  author={Mehr, Goodarz and Gohari, Sepideh and Abbas, Montasir and Eskandarian, Azim},
  journal={arXiv preprint arXiv:2607.23910},
  year={2026}
}

@inproceedings{yang2026mmsi,
  title={Mmsi-bench: A benchmark for multi-image spatial intelligence},
  author={Yang, Sihan and Xu, Runsen and Xie, Yiman and Yang, Sizhe and Li, Mo and Lin, Jingli and Zhu, Chenming and Chen, Xiaochen and Duan, Haodong and Yue, Xiangyu and others},
  booktitle={International Conference on Learning Representations},
  volume={2026},
  pages={157051--157088},
  year={2026}
}

@inproceedings{yeh2026seeing,
  title={Seeing from another perspective: Evaluating multi-view understanding in mllms},
  author={Yeh, Chun-Hsiao and Wang, Chenyu and Tong, Shengbang and Cheng, Ta-Ying and Wang, Ruoyu and Chu, Tianzhe and Zhai, Yuexiang and Chen, Yubei and Gao, Shenghua and Ma, Yi},
  booktitle={Proceedings of the AAAI Conference on Artificial Intelligence},
  volume={40},
  number={14},
  pages={12000--12008},
  year={2026}
}

@article{zhang2026flatland,
  title={From flatland to space: Teaching vision-language models to perceive and reason in 3d},
  author={Zhang, Jiahui and Chen, Yurui and Xu, Yueming and Huang, Ze and Mei, Jilin and Chen, Chunhui and Zhou, Yanpeng and Yuan, Yu-Jie and Cai, Xinyue and Huang, Guowei and others},
  journal={Advances in Neural Information Processing Systems},
  volume={38},
  year={2026}
}

@article{feng2025seeing,
  title={Seeing across views: Benchmarking spatial reasoning of vision-language models in robotic scenes},
  author={Feng, Zhiyuan and Kang, Zhaolu and Wang, Qijie and Du, Zhiying and Yan, Jiongrui and Shi, Shubin and Yuan, Chengbo and Liang, Huizhi and Deng, Yu and Li, Qixiu and others},
  journal={arXiv preprint arXiv:2510.19400},
  year={2025}
}

@inproceedings{xu2026citycube,
  title={Citycube: Benchmarking cross-view spatial reasoning on vision-language models in urban environments},
  author={Xu, Haotian and Hu, Yue and Zhu, Zhengqiu and Gao, Chen and Wang, Ziyou and Rao, Junreng and Lu, Wenhao and Li, Weishi and Yin, Quanjun and Li, Yong},
  booktitle={Proceedings of the 64th Annual Meeting of the Association for Computational Linguistics (Volume 1: Long Papers)},
  pages={9190--9215},
  year={2026}
}

@inproceedings{jeong2026learning,
  title={Learning Multi-View Spatial Reasoning from Cross-View Relations},
  author={Jeong, Suchae and Song, Jaehwi and Lee, Haeone and Kim, Hanna and Kim, Jian and Lee, Dongjun and Shin, Dong Kyu and Kim, Changyeon and Hahm, Dongyoon and Jin, Woogyeol and others},
  booktitle={Proceedings of the IEEE/CVF Conference on Computer Vision and Pattern Recognition},
  pages={2570--2581},
  year={2026}
}

@inproceedings{zimmer2024tumtraf,
  title={Tumtraf v2x cooperative perception dataset},
  author={Zimmer, Walter and Wardana, Gerhard Arya and Sritharan, Suren and Zhou, Xingcheng and Song, Rui and Knoll, Alois C},
  booktitle={2024 IEEE/CVF Conference on Computer Vision and Pattern Recognition (CVPR)},
  pages={22668--22677},
  year={2024},
  organization={IEEE}
}

@book{lewis69,
  address = {Cambridge, MA, USA},
  author = {Lewis, Dk},
  edition = {First},
  publisher = {Harvard University Press},
  title = {Convention: a philosophical study},
  year = 1969
}

@article{lazaridou2016multi,
  title={Multi-agent cooperation and the emergence of (natural) language},
  author={Lazaridou, Angeliki and Peysakhovich, Alexander and Baroni, Marco},
  journal={arXiv preprint arXiv:1612.07182},
  year={2016}
}

@article{foerster2016learning,
  title={Learning to communicate with deep multi-agent reinforcement learning},
  author={Foerster, Jakob and Assael, Ioannis Alexandros and De Freitas, Nando and Whiteson, Shimon},
  journal={Advances in neural information processing systems},
  volume={29},
  year={2016}
}

@inproceedings{foerster2018counterfactual,
  title={Counterfactual multi-agent policy gradients},
  author={Foerster, Jakob and Farquhar, Gregory and Afouras, Triantafyllos and Nardelli, Nantas and Whiteson, Shimon},
  booktitle={Proceedings of the AAAI conference on artificial intelligence},
  volume={32},
  number={1},
  year={2018}
}

@article{rashid2020monotonic,
  title={Monotonic value function factorisation for deep multi-agent reinforcement learning},
  author={Rashid, Tabish and Samvelyan, Mikayel and De Witt, Christian Schroeder and Farquhar, Gregory and Foerster, Jakob and Whiteson, Shimon},
  journal={Journal of Machine Learning Research},
  volume={21},
  number={178},
  pages={1--51},
  year={2020}
}

@article{shao2024deepseekmath,
  title={Deepseekmath: Pushing the limits of mathematical reasoning in open language models},
  author={Shao, Zhihong and Wang, Peiyi and Zhu, Qihao and Xu, Runxin and Song, Junxiao and Bi, Xiao and Zhang, Haowei and Zhang, Mingchuan and Li, YK and Wu, Yang and others},
  journal={arXiv preprint arXiv:2402.03300},
  year={2024}
}

@article{liu2025understanding,
  title={Understanding r1-zero-like training: A critical perspective},
  author={Liu, Zichen and Chen, Changyu and Li, Wenjun and Qi, Penghui and Pang, Tianyu and Du, Chao and Lee, Wee Sun and Lin, Min},
  journal={arXiv preprint arXiv:2503.20783},
  year={2025}
}

@article{Zheng2025GroupSP,
  title={Group Sequence Policy Optimization},
  author={Chujie Zheng and Shi-Xuan Liu and Ming-Ze Li and Xiong-Hui Chen and Bo-Wen Yu and Chang Gao and Kai Dang and Yu-Qiong Liu and Rui Men and An Yang and Jing-Ren Zhou and Jun-Yang Lin},
  journal={ArXiv},
  year={2025},
  volume={abs/2507.18071},
  url={https://api.semanticscholar.org/CorpusID:280017753}
}

@article{chen2026exact,
  title={Exact is easier: Credit assignment for cooperative llm agents},
  author={Chen, Yanjun and Sun, Yirong and Wang, Hanlin and Wang, Jinghan and Zhang, Xinming and Shen, Xiaoyu and Li, Wenjie and Zhang, Wei},
  journal={arXiv preprint arXiv:2603.06859},
  year={2026}
}

@article{li2026counterfactual,
  title={Counterfactual Credit Policy Optimization for Multi-Agent Collaboration},
  author={Li, Zhongyi and Tian, Wan and Chen, Jinju and Zhang, Huiming and Liu, Yang and Ban, Yikun and Zhuang, Fuzhen},
  journal={arXiv preprint arXiv:2603.21563},
  year={2026}
}

@article{hu2021lora,
  title={Lora: Low-rank adaptation of large language models},
  author={Hu, Edward J and Shen, Yelong and Wallis, Phillip and Allen-Zhu, Zeyuan and Li, Yuanzhi and Wang, Shean and Wang, Lu and Chen, Weizhu},
  journal={arXiv preprint arXiv:2106.09685},
  year={2021}
}

@article{bai2025qwen3,
  title={Qwen3-vl technical report},
  author={Bai, Shuai and Cai, Yuxuan and Chen, Ruizhe and Chen, Keqin and Chen, Xionghui and Cheng, Zesen and Deng, Lianghao and Ding, Wei and Gao, Chang and Ge, Chunjiang and others},
  journal={arXiv preprint arXiv:2511.21631},
  year={2025}
}

@article{wang2025internvl3,
  title={Internvl3. 5: Advancing open-source multimodal models in versatility, reasoning, and efficiency},
  author={Wang, Weiyun and Gao, Zhangwei and Gu, Lixin and Pu, Hengjun and Cui, Long and Wei, Xingguang and Liu, Zhaoyang and Jing, Linglin and Ye, Shenglong and Shao, Jie and others},
  journal={arXiv preprint arXiv:2508.18265},
  year={2025}
}

@article{li2024llava,
  title={Llava-onevision: Easy visual task transfer},
  author={Li, Bo and Zhang, Yuanhan and Guo, Dong and Zhang, Renrui and Li, Feng and Zhang, Hao and Zhang, Kaichen and Zhang, Peiyuan and Li, Yanwei and Liu, Ziwei and others},
  journal={arXiv preprint arXiv:2408.03326},
  year={2024}
}

@article{agrawal2024pixtral,
  title={Pixtral 12B},
  author={Agrawal, Pravesh and Antoniak, Szymon and Hanna, Emma Bou and Bout, Baptiste and Chaplot, Devendra and Chudnovsky, Jessica and Costa, Diogo and De Monicault, Baudouin and Garg, Saurabh and Gervet, Theophile and others},
  journal={arXiv preprint arXiv:2410.07073},
  year={2024}
}

@article{team2025gemma,
  title={Gemma 3 technical report},
  author={Team, Gemma and Kamath, Aishwarya and Ferret, Johan and Pathak, Shreya and Vieillard, Nino and Merhej, Ramona and Perrin, Sarah and Matejovicova, Tatiana and Ram{\'e}, Alexandre and Rivi{\`e}re, Morgane and others},
  journal={arXiv preprint arXiv:2503.19786},
  year={2025}
}

@inproceedings{fu2024blink,
  title={Blink: Multimodal large language models can see but not perceive},
  author={Fu, Xingyu and Hu, Yushi and Li, Bangzheng and Feng, Yu and Wang, Haoyu and Lin, Xudong and Roth, Dan and Smith, Noah A and Ma, Wei-Chiu and Krishna, Ranjay},
  booktitle={European Conference on Computer Vision},
  pages={148--166},
  year={2024},
  organization={Springer}
}

@inproceedings{yang2025thinking,
  title={Thinking in space: How multimodal large language models see, remember, and recall spaces},
  author={Yang, Jihan and Yang, Shusheng and Gupta, Anjali W and Han, Rilyn and Fei-Fei, Li and Xie, Saining},
  booktitle={2025 IEEE/CVF Conference on Computer Vision and Pattern Recognition (CVPR)},
  pages={10632--10643},
  year={2025},
  organization={IEEE}
}

@inproceedings{wang2025spatial,
  title={Spatial mental modeling from limited views},
  author={Wang, Qineng and Yin, Baiqiao and Zhang, Pingyue and Zhang, Jianshu and Wang, Kangrui and Wang, Zihan and Zhang, Jieyu and Chandrasegaran, Keshigeyan and Liu, Han and Krishna, Ranjay and others},
  booktitle={The Fourteenth International Conference on Learning Representations},
  volume={4},
  year={2025}
}

@InProceedings{Chen_2024_CVPR,
    author    = {Chen, Boyuan and Xu, Zhuo and Kirmani, Sean and Ichter, Brain and Sadigh, Dorsa and Guibas, Leonidas and Xia, Fei},
    title     = {SpatialVLM: Endowing Vision-Language Models with Spatial Reasoning Capabilities},
    booktitle = {Proceedings of the IEEE/CVF Conference on Computer Vision and Pattern Recognition (CVPR)},
    month     = {June},
    year      = {2024},
    pages     = {14455-14465}
}

@article{cheng2024spatialrgpt,
  title={Spatialrgpt: Grounded spatial reasoning in vision-language models},
  author={Cheng, An-Chieh and Yin, Hongxu and Fu, Yang and Guo, Qiushan and Yang, Ruihan and Kautz, Jan and Wang, Xiaolong and Liu, Sifei},
  journal={Advances in Neural Information Processing Systems},
  volume={37},
  pages={135062--135093},
  year={2024}
}

@inproceedings{ma20253dsrbench,
  title={3dsrbench: A comprehensive 3d spatial reasoning benchmark},
  author={Ma, Wufei and Chen, Haoyu and Zhang, Guofeng and Chou, Yu-Cheng and Chen, Jieneng and De Melo, Celso and Yuille, Alan},
  booktitle={2025 IEEE/CVF International Conference on Computer Vision (ICCV)},
  pages={6924--6934},
  year={2025},
  organization={IEEE}
}

@article{tong2024cambrian,
  title={Cambrian-1: A fully open, vision-centric exploration of multimodal llms},
  author={Tong, Shengbang and Brown, Ellis and Wu, Penghao and Woo, Sanghyun and Middepogu, Manoj and Akula, Sai C and Yang, Jihan and Yang, Shusheng and Iyer, Adithya and Pan, Xichen and others},
  journal={Advances in Neural Information Processing Systems},
  volume={37},
  pages={87310--87356},
  year={2024}
}

@article{li2022v2x,
  title={V2X-Sim: Multi-agent collaborative perception dataset and benchmark for autonomous driving},
  author={Li, Yiming and Ma, Dekun and An, Ziyan and Wang, Zixun and Zhong, Yiqi and Chen, Siheng and Feng, Chen},
  journal={IEEE robotics and automation letters},
  volume={7},
  number={4},
  pages={10914--10921},
  year={2022},
  publisher={IEEE}
}

@InProceedings{Ma_2024_WACV,
    author    = {Ma, Yunsheng and Lu, Juanwu and Cui, Can and Zhao, Sicheng and Cao, Xu and Ye, Wenqian and Wang, Ziran},
    title     = {MACP: Efficient Model Adaptation for Cooperative Perception},
    booktitle = {Proceedings of the IEEE/CVF Winter Conference on Applications of Computer Vision (WACV)},
    month     = {January},
    year      = {2024},
    pages     = {3373-3382}
}

@inproceedings{xu2022opv2v,
  title={Opv2v: An open benchmark dataset and fusion pipeline for perception with vehicle-to-vehicle communication},
  author={Xu, Runsheng and Xiang, Hao and Xia, Xin and Han, Xu and Li, Jinlong and Ma, Jiaqi},
  booktitle={2022 International Conference on Robotics and Automation (ICRA)},
  pages={2583--2589},
  year={2022},
  organization={IEEE}
}

@inproceedings{yu2022dair,
  title={Dair-v2x: A large-scale dataset for vehicle-infrastructure cooperative 3d object detection},
  author={Yu, Haibao and Luo, Yizhen and Shu, Mao and Huo, Yiyi and Yang, Zebang and Shi, Yifeng and Guo, Zhenglong and Li, Hanyu and Hu, Xing and Yuan, Jirui and others},
  booktitle={2022 IEEE/CVF Conference on Computer Vision and Pattern Recognition (CVPR)},
  pages={21329--21338},
  year={2022},
  organization={IEEE}
}

@article{xu2022cobevt,
  title={CoBEVT: Cooperative bird's eye view semantic segmentation with sparse transformers},
  author={Xu, Runsheng and Tu, Zhengzhong and Xiang, Hao and Shao, Wei and Zhou, Bolei and Ma, Jiaqi},
  journal={arXiv preprint arXiv:2207.02202},
  year={2022}
}

@article{hu2022where2comm,
  title={Where2comm: Communication-efficient collaborative perception via spatial confidence maps},
  author={Hu, Yue and Fang, Shaoheng and Lei, Zixing and Zhong, Yiqi and Chen, Siheng},
  journal={Advances in neural information processing systems},
  volume={35},
  pages={4874--4886},
  year={2022}
}

@article{havrylov2017emergence,
  title={Emergence of language with multi-agent games: Learning to communicate with sequences of symbols},
  author={Havrylov, Serhii and Titov, Ivan},
  journal={Advances in neural information processing systems},
  volume={30},
  year={2017}
}

@article{chaabouni2019anti,
  title={Anti-efficient encoding in emergent communication},
  author={Chaabouni, Rahma and Kharitonov, Eugene and Dupoux, Emmanuel and Baroni, Marco},
  journal={Advances in neural information processing systems},
  volume={32},
  year={2019}
}

@article{lowe2019pitfalls,
  title={On the pitfalls of measuring emergent communication},
  author={Lowe, Ryan and Foerster, Jakob and Boureau, Y-Lan and Pineau, Joelle and Dauphin, Yann},
  journal={arXiv preprint arXiv:1903.05168},
  year={2019}
}

@article{sunehag2017value,
  title={Value-decomposition networks for cooperative multi-agent learning},
  author={Sunehag, Peter and Lever, Guy and Gruslys, Audrunas and Czarnecki, Wojciech Marian and Zambaldi, Vinicius and Jaderberg, Max and Lanctot, Marc and Sonnerat, Nicolas and Leibo, Joel Z and Tuyls, Karl and others},
  journal={arXiv preprint arXiv:1706.05296},
  year={2017}
}

@article{lowe2017multi,
  title={Multi-agent actor-critic for mixed cooperative-competitive environments},
  author={Lowe, Ryan and Wu, Yi I and Tamar, Aviv and Harb, Jean and Pieter Abbeel, OpenAI and Mordatch, Igor},
  journal={Advances in neural information processing systems},
  volume={30},
  year={2017}
}

@article{yu2022surprising,
  title={The surprising effectiveness of ppo in cooperative multi-agent games},
  author={Yu, Chao and Velu, Akash and Vinitsky, Eugene and Gao, Jiaxuan and Wang, Yu and Bayen, Alexandre and Wu, Yi},
  journal={Advances in neural information processing systems},
  volume={35},
  pages={24611--24624},
  year={2022}
}

@article{du2023improving,
  title={Improving factuality and reasoning in language models through multiagent debate},
  author={Du, Yilun and Li, Shuang and Torralba, Antonio and Tenenbaum, Joshua B and Mordatch, Igor},
  journal={arXiv preprint arXiv:2305.14325},
  year={2023}
}

@article{li2023camel,
  title={Camel: Communicative agents for" mind" exploration of large language model society},
  author={Li, Guohao and Hammoud, Hasan and Itani, Hani and Khizbullin, Dmitrii and Ghanem, Bernard},
  journal={Advances in neural information processing systems},
  volume={36},
  pages={51991--52008},
  year={2023}
}

@article{wu2023autogen,
  title={Autogen: Enabling next-gen llm applications via multi-agent conversation},
  author={Wu, Qingyun and Bansal, Gagan and Zhang, Jieyu and Wu, Yiran and Li, Beibin and Zhu, Erkang and Jiang, Li and Zhang, Xiaoyun and Zhang, Shaokun and Liu, Jiale and others},
  journal={arXiv preprint arXiv:2308.08155},
  year={2023}
}

@article{Schulman2017ProximalPO,
  title={Proximal Policy Optimization Algorithms},
  author={John D. Schulman and Filip Wolski and Prafulla Dhariwal and Alec Radford and Oleg Klimov},
  journal={ArXiv},
  year={2017},
  volume={abs/1707.06347},
  url={https://api.semanticscholar.org/CorpusID:28695052}
}

@inproceedings{ahmadian-etal-2024-back,
    title = "Back to Basics: Revisiting {REINFORCE}-Style Optimization for Learning from Human Feedback in {LLM}s",
    author = {Ahmadian, Arash  and
      Cremer, Chris  and
      Gall{\'e}, Matthias  and
      Fadaee, Marzieh  and
      Kreutzer, Julia  and
      Pietquin, Olivier  and
      {\"U}st{\"u}n, Ahmet  and
      Hooker, Sara},
    editor = "Ku, Lun-Wei  and
      Martins, Andre  and
      Srikumar, Vivek",
    booktitle = "Proceedings of the 62nd Annual Meeting of the {A}ssociation for {C}omputational {L}inguistics (Volume 1: Long Papers)",
    month = aug,
    year = "2024",
    address = "Bangkok, Thailand",
    publisher = "Association for Computational Linguistics",
    url = "https://aclanthology.org/2024.acl-long.662/",
    doi = "10.18653/v1/2024.acl-long.662",
    pages = "12248--12267"
}

@article{guo2025deepseek,
  title={Deepseek-r1: Incentivizing reasoning capability in llms via reinforcement learning},
  author={Guo, Daya and Yang, Dejian and Zhang, Haowei and Song, Junxiao and Wang, Peiyi and Zhu, Qihao and Xu, Runxin and Zhang, Ruoyu and Ma, Shirong and Bi, Xiao and others},
  journal={arXiv preprint arXiv:2501.12948},
  year={2025}
}

@article{liao2025improved,
  title={Improved visual-spatial reasoning via r1-zero-like training},
  author={Liao, Zhenyi and Xie, Qingsong and Zhang, Yanhao and Kong, Zijian and Lu, Haonan and Yang, Zhenyu and Deng, Zhijie},
  journal={arXiv preprint arXiv:2504.00883},
  year={2025}
}

@article{cho2026spatialclaw,
  title={SpatialClaw: Rethinking Action Interface for Agentic Spatial Reasoning},
  author={Cho, Seokju and Hachiuma, Ryo and Badki, Abhishek and Su, Hang and Lee, Byung-Kwan and Song, Chan Hee and Liu, Sifei and Radhakrishnan, Subhashree and Kim, Seungryong and Wang, Yu-Chiang Frank and others},
  journal={arXiv preprint arXiv:2606.13673},
  year={2026}
}

@inproceedings{li2026spatialladder,
  title={Spatialladder: Progressive training for spatial reasoning in vision-language models},
  author={Li, Hongxing and Li, Dingming and Wang, Zixuan and Yan, Yuchen and Wu, Hang and Zhang, Wenqi and Shen, Yongliang and Lu, Weiming and Xiao, Jun and Zhuang, Yueting},
  booktitle={International Conference on Learning Representations},
  volume={2026},
  pages={76566--76592},
  year={2026}
}

@article{gong2026s,
  title={S-Agent 2.0: Spatial Reasoning with a General Coding Agent},
  author={Gong, Rui and Wang, Ruoyu and Dai, Yalun and Li, Hao and Yang, Sidi and Zhao, Qihao and Li, Hongyu and Guan, Zeyang and Hong, Fangzhou and Chen, Zhaoxi and others},
  journal={arXiv preprint arXiv:2610.28354},
  year={2026}
}

@article{yu2026evospatial,
  title={EvoSpatial: Experience-Driven Skill and Tool Evolution for Spatial Reasoning Agents},
  author={Yu, Bin and Lian, Shijie and Shen, Zhaolong and Lin, Xiaopeng and Zhang, Yao and Liu, Haishan and Li, Haotian and Wei, Yuliang and Hu, Kai and Du, Yichao and others},
  journal={arXiv preprint arXiv:2610.16679},
  year={2026}
}

@inproceedings{mandi2024roco,
  title={Roco: Dialectic multi-robot collaboration with large language models},
  author={Mandi, Zhao and Jain, Shreeya and Song, Shuran},
  booktitle={2024 IEEE International Conference on Robotics and Automation (ICRA)},
  pages={286--299},
  year={2024},
  organization={IEEE}
}

@inproceedings{zhang2024building,
  title={Building cooperative embodied agents modularly with large language models},
  author={Zhang, Hongxin and Du, Weihua and Shan, Jiaming and Zhou, Qinhong and Du, Yilun and Tenenbaum, Joshua B and Shu, Tianmin and Gan, Chuang},
  booktitle={International Conference on Learning Representations},
  volume={2024},
  pages={19373--19401},
  year={2024}
}

@inproceedings{chen2024scalable,
  title={Scalable multi-robot collaboration with large language models: Centralized or decentralized systems?},
  author={Chen, Yongchao and Arkin, Jacob and Zhang, Yang and Roy, Nicholas and Fan, Chuchu},
  booktitle={2024 IEEE International Conference on Robotics and Automation (ICRA)},
  pages={4311--4317},
  year={2024},
  organization={IEEE}
}

@inproceedings{2900423.2900661,
author = {Tellex, Stefanie and Kollar, Thomas and Dickerson, Steven and Walter, Matthew R. and Banerjee, Ashis Gopal and Teller, Seth and Roy, Nicholas},
title = {Understanding natural language commands for robotic navigation and mobile manipulation},
year = {2011},
publisher = {AAAI Press},
booktitle = {Proceedings of the Twenty-Fifth AAAI Conference on Artificial Intelligence},
pages = {1507–1514},
numpages = {8},
location = {San Francisco, California},
series = {AAAI'11}
}

@inproceedings{park2023generative,
  title={Generative agents: Interactive simulacra of human behavior},
  author={Park, Joon Sung and O'Brien, Joseph and Cai, Carrie Jun and Morris, Meredith Ringel and Liang, Percy and Bernstein, Michael S},
  booktitle={Proceedings of the 36th annual acm symposium on user interface software and technology},
  pages={1--22},
  year={2023}
}

@inproceedings{lightman2024let,
  title={Let's verify step by step},
  author={Lightman, Hunter and Kosaraju, Vineet and Burda, Yuri and Edwards, Harrison and Baker, Bowen and Lee, Teddy and Leike, Jan and Schulman, John and Sutskever, Ilya and Cobbe, Karl},
  booktitle={International Conference on Learning Representations},
  volume={2024},
  pages={39578--39601},
  year={2024}
}

@inproceedings{li2024improving,
  title={Improving multi-agent debate with sparse communication topology},
  author={Li, Yunxuan and Du, Yibing and Zhang, Jiageng and Hou, Le and Grabowski, Peter and Li, Yeqing and Ie, Eugene},
  booktitle={Findings of the Association for Computational Linguistics: EMNLP 2024},
  pages={7281--7294},
  year={2024}
}

@inproceedings{feng2024don,
  title={Don’t hallucinate, abstain: Identifying llm knowledge gaps via multi-llm collaboration},
  author={Feng, Shangbin and Shi, Weijia and Wang, Yike and Ding, Wenxuan and Balachandran, Vidhisha and Tsvetkov, Yulia},
  booktitle={Proceedings of the 62nd Annual Meeting of the Association for Computational Linguistics (Volume 1: Long Papers)},
  pages={14664--14690},
  year={2024}
}

@inproceedings{wu2024v,
  title={V*: Guided visual search as a core mechanism in multimodal llms},
  author={Wu, Penghao and Xie, Saining},
  booktitle={2024 IEEE/CVF Conference on Computer Vision and Pattern Recognition (CVPR)},
  pages={13084--13094},
  year={2024},
  organization={IEEE}
}

@inproceedings{chen2024reconcile,
  title={Reconcile: Round-table conference improves reasoning via consensus among diverse llms},
  author={Chen, Justin and Saha, Swarnadeep and Bansal, Mohit},
  booktitle={Proceedings of the 62nd Annual Meeting of the Association for Computational Linguistics (Volume 1: Long Papers)},
  pages={7066--7085},
  year={2024}
}

@inproceedings{hong2024metagpt,
  title={MetaGPT: Meta programming for a multi-agent collaborative framework},
  author={Hong, Sirui and Zhuge, Mingchen and Chen, Jonathan and Zheng, Xiawu and Cheng, Yuheng and Wang, Jinlin and Zhang, Ceyao and Yau, Steven and Lin, Zijuan and Zhou, Liyang and others},
  booktitle={International Conference on Learning Representations},
  volume={2024},
  pages={23247--23275},
  year={2024}
}

@inproceedings{das2017learning,
  title={Learning cooperative visual dialog agents with deep reinforcement learning},
  author={Das, Abhishek and Kottur, Satwik and Moura, Jos{\'e} MF and Lee, Stefan and Batra, Dhruv},
  booktitle={2017 IEEE International conference on computer vision (ICCV)},
  pages={2970--2979},
  year={2017},
  organization={IEEE}
}

@article{kojima2022large,
  title={Large language models are zero-shot reasoners},
  author={Kojima, Takeshi and Gu, Shixiang Shane and Reid, Machel and Matsuo, Yutaka and Iwasawa, Yusuke},
  journal={Advances in neural information processing systems},
  volume={35},
  pages={22199--22213},
  year={2022}
}
}

\maketitlesupplementary
\appendix
\crefalias{section}{appendix}
\setcounter{section}{0}
\renewcommand{\thesection}{\Alph{section}}
\setcounter{table}{0}
\renewcommand{\thetable}{S\arabic{table}}

\noindent\textbf{Contents}
\vspace{2pt}
\begin{itemize}\setlength\itemsep{0pt}\setlength\parskip{0pt}\small
  \item[\textbf{\ref{sec:supp:prompts}}] \nameref{sec:supp:prompts} \dotfill \pageref{sec:supp:prompts}
  \item[\textbf{\ref{sec:supp:setup}}] \nameref{sec:supp:setup} \dotfill \pageref{sec:supp:setup}
  \item[\textbf{\ref{sec:supp:compute}}] \nameref{sec:supp:compute} \dotfill \pageref{sec:supp:compute}
  \item[\textbf{\ref{sec:supp:ceiling}}] \nameref{sec:supp:ceiling} \dotfill \pageref{sec:supp:ceiling}
  \item[\textbf{\ref{sec:supp:rl}}] \nameref{sec:supp:rl} \dotfill \pageref{sec:supp:rl}
  \item[\textbf{\ref{sec:supp:transfer}}] \nameref{sec:supp:transfer} \dotfill \pageref{sec:supp:transfer}
  \item[\textbf{\ref{sec:supp:designs}}] \nameref{sec:supp:designs} \dotfill \pageref{sec:supp:designs}
  \item[\textbf{\ref{sec:supp:sft}}] \nameref{sec:supp:sft} \dotfill \pageref{sec:supp:sft}
  \item[\textbf{\ref{sec:supp:freeride}}] \nameref{sec:supp:freeride} \dotfill \pageref{sec:supp:freeride}
  \item[\textbf{\ref{sec:supp:qual}}] \nameref{sec:supp:qual} \dotfill \pageref{sec:supp:qual}
\end{itemize}
\vspace{4pt}

\vspace{4pt}

\section{Prompt suites}
\label{sec:supp:prompts}
A few prompt details are given below; braces mark substituted fields, $B$ is the word
budget and $w$ the listener's reasoning cap. Colour indicates the role, speaker, blind
listener, scoring listener and centralised agent.

\begin{promptbox}[pbSpeak]{Speaker (decentralised, all designs)}
\pbt{You are one of \{n\} observers of a scene. You can see\\
only Image \{i\}; every other view is hidden from you.\\
Another agent, who sees no image at all, must answer\\
the question below using only what you and the other\\
observers report.\\
~\\
Question: \{q\}\\
~\\
In at most \{B\} words, describe exactly what is\\
visible in YOUR view that bears on this question:\\
the objects, where they are, and their spatial\\
relations. State only what you can see. Do not guess\\
the answer.}
\end{promptbox}
Each speaker receives $N$ image slots of which only its own carries the real view, the rest
holding grey canvases of identical size. It is given the question but never the options.

\begin{promptbox}[pbList]{Blind listener, stage 1 (reasoning, no images)}
\pbt{\{n\} observers each saw one view of a scene. You\\
cannot see any view. They reported:\\
  Observer 1: \{u\_1\}\\
  ...\\
  Observer \{n\}: \{u\_N\}\\
~\\
Reason step by step about what the combined reports\\
imply about the scene: work out which observers are\\
describing the same objects, reconcile their\\
viewpoints, and only then decide. Think in at most\\
\{w\} words.\\
Question: \{q\}\\
Your reasoning:}
\end{promptbox}
\begin{promptbox}[pbList]{Blind listener, stage 2 (answer, letters scored)}
\pbt{<the reasoning prompt above, verbatim>\\
~\\
Your reasoning: \{think\}\\
~\\
Question: \{q\}\\
\{options\}\\
~\\
Answer with a single letter (\{letters\}). Answer:}
\end{promptbox}
\begin{promptbox}[pbScore]{Scoring listener (one call, no reasoning)}
\pbt{\{n\} observers each saw one view of a scene. You\\
cannot see any of the views. Each reported what they\\
could see.\\
~\\
Observer 1 (Image 1): \{u\_1\}\\
...\\
Observer \{n\} (Image \{n\}): \{u\_N\}\\
~\\
Question: \{q\}\\
\{options\}\\
~\\
Answer with a single letter (\{letters\}). Answer:}
\end{promptbox}
This is the decoder that defines the reward of Eq.~(\ref{eq:reward}). Its wording differs from
the chain-of-thought listener rather than being a truncation of it.

\paragraph{Why the listener reasons in two stages.} The chain-of-thought listener is called
twice, once to produce reasoning from the observers' reports and once to score the option
letters with that reasoning appended. This is the standard two-stage form of zero-shot
chain-of-thought prompting, in which a first call elicits the rationale and a second call
extracts the answer~\cite{kojima2022large}. Scoring the letters rather than generating one is also
what Eq.~(\ref{eq:reward}) requires, since the reward is the probability the listener assigns to
the gold option and a free-text answer exposes no such quantity.

The one departure from common practice is that stage one does not show the options. This is
deliberate. A listener that reasons over the options can eliminate distractors on
plausibility alone and still score well on messages carrying no view-specific content,
which is the effect the A/B/C partition exists to exclude. We measured what the choice
costs by re-scoring stored messages from the joint-credit policy of Table~\ref{tab:rl} under both
stage-one prompts, holding everything else fixed: same frozen listener, same greedy
decoding, same $969$ questions.

\begin{table}[h]
  \centering\small
  \setlength{\tabcolsep}{5pt}
  \begin{tabular}{@{}lccccc@{}}
    \toprule
    stage-one prompt & ALL & A & B & C & B\,$+$\,C \\
    \midrule
    options hidden (ours) & 38.7 & 47.9 & 44.9 & 30.3 & 34.7 \\
    options shown         & 39.5 & 51.4 & 46.8 & 29.0 & 34.4 \\
    \bottomrule
  \end{tabular}
  \caption{Listener prompt ablation. Messages from one trained policy re-scored under both
  stage-one prompts by the same frozen listener; $969$ questions, $677$ on B\,$+$\,C. Only
  the visibility of the options changes.}
  \label{tab:supp:listener}
\end{table}

Hiding the options costs nothing on B+C, where it is $0.3$ points ahead, so the design does
not handicap the listener. The informative part is that the strata move in opposite
directions. Accuracy rises on A and B, the slices where option structure is partly
exploitable without the messages, and falls on C, the slice that requires reconciling every
view: reasoning anchored on the options drifts toward plausible distractors rather than the
observers' reports. The parity on B+C is therefore a near-cancellation of two opposing
effects and not an equivalence between the prompts, which the paired disagreement makes
plain, since the two prompts return different answers on $42\%$ of B+C questions.

\begin{promptbox}[pbCent]{Centralised reference, chain-of-thought variant}
\pbt{You can see all \{n\} views of one scene, shown as\\
Image 1 to Image \{n\}.\\
Reason step by step about what the views jointly\\
imply about the scene's layout, then decide. Think\\
in at most \{w\} words.\\
Question: \{q\}\\
Your reasoning:}
\end{promptbox}
\begin{promptbox}[pbCent]{Centralised reference, scoring variant}
\pbt{You can see all \{n\} views of one scene, shown as\\
Image 1 to Image \{n\}.\\
~\\
Question: \{q\}\\
\{options\}\\
~\\
Answer with a single letter (\{letters\}). Answer:}
\end{promptbox}
Both centralised calls receive all $N$ real views. There is no speaker stage and no message
channel, so the centralised arm is a single agent rather than a sighted listener.

Table~\ref{tab:designs} singles out
\textsc{p2p} as the best admissible design and \textsc{abstain} as the clear failure, so we
give those two prompts in full; the remaining designs vary the speaker prompt, the message
routing or the number of rounds around the same template. \textsc{p2p} runs two speaker
calls. Each agent first writes one question for a peer, the questions are routed so that
agent $i$ answers the question written by agent $i-1$, and the message finally sent is the
agent's own description concatenated with its answer to the peer.

\begin{promptbox}[pbSpeak]{\textsc{p2p}, call 1 (each agent writes one question)}
\pbt{You are one of \{n\} observers, seeing only Image \{i\}.\\
To help answer the question below you may ask ONE\\
other observer one question, at most \{B\} words.\\
Output only the question.\\
~\\
Question: \{q\}\\
~\\
Your question:}
\end{promptbox}

\begin{promptbox}[pbSpeak]{\textsc{p2p}, call 2 (each agent answers its peer)}
\pbt{You are one of \{n\} observers, seeing only Image \{i\}.\\
Another agent asked:\\
"\{ask\}"\\
Answer in at most \{B\} words using only what is\\
visible in YOUR view.}
\end{promptbox}

\begin{promptbox}[pbSpeak]{\textsc{abstain} (speaker may decline)}
\pbt{In at most \{B\} words describe what in YOUR view\\
bears on the question. If your view shows nothing\\
relevant, reply exactly: NOTHING RELEVANT.}
\end{promptbox}

A speaker that replies \texttt{NOTHING RELEVANT} is dropped from the message set, and if
every speaker declines the full set is retained so that the listener always receives
something.

\section{Experimental setting}
\label{sec:supp:setup}
Speakers decode greedily, so inference is deterministic and
repeated runs on the same question reproduce bit-identically. The chain-of-thought listener
is capped at $512$ words of reasoning, realised as a token ceiling with headroom so that
generations are not truncated mid-sentence. Views hidden from a speaker are replaced by
grey canvases of the same dimensions rather than removed, which keeps the number of image
tokens constant across configurations so that no measured gap can arise from differing
context length. All images pass through a shared token budget that rescales them jointly,
so the centralised arm and the decentralised arm see the same pixels at the same
resolution. Because the blind gate measures exploitation of option structure, option order
is shuffled once with a fixed seed and the permutation recorded, so the letter-prior
correction is applied against a pool with no positional signal.

\section{Computational resources}
\label{sec:supp:compute}
All experiments were run on GH200 nodes. Each node
carries one NVIDIA H200 GPU with $96$\,GiB of HBM3 memory alongside a $72$-core NVIDIA Grace
CPU with $120$\,GiB of LPDDR5X, and jobs are charged at one node-hour per node-hour. We used
$8$ nodes per training or inference job. The project consumed approximately $4{,}000$ GH200
node-hours end to end.
The broad split is as follows. Roughly two thirds is inference: the five-backbone,
two-certifier gap matrix, the $27$-design matrix, the oracle ceiling, and the evaluation
sweeps that score every trained policy on both certified test sets, in places under both
listeners. Roughly one third is reinforcement-learning training. The
remainder, a few percent, covers benchmark construction and the two certification passes.
Training updates LoRA adapters only, so the memory footprint is dominated by the frozen
backbone and the $K$ sampled rollouts per question.

\section{The oracle ceiling}
\label{sec:supp:ceiling}
The ceiling row of Table~\ref{tab:rl} is obtained by sampling $K{=}8$ message sets per question
and keeping, per question, the set on which the listener assigns the highest probability to
the gold answer. Selection therefore consults the answer, which is what makes the row an
upper bound on what can be transmitted at this budget rather than an accuracy any policy
could reach unaided. The per-stratum values bound what selection over eight transcripts
recovers, not what a single rollout does.

\section{Reinforcement learning}
\label{sec:supp:rl}

\paragraph{Settings.} Every variant shares one configuration: $K{=}4$ sampled message sets
per question, sampling temperature $0.9$, KL coefficient $\beta{=}0.02$, learning rate
$2\times10^{-5}$, and a LoRA adapter of rank $64$ and $\alpha{=}128$ shared across all $N$
speaker roles. Sequence-level variants take two inner epochs per generated batch with
GSPO's asymmetric clip bounds $\varepsilon_{\text{lo}}{=}3\times10^{-4}$ and
$\varepsilon_{\text{hi}}{=}4\times10^{-4}$. Only the credit term of
Eqs.~(\ref{eq:joint}), (\ref{eq:loo}) and~(\ref{eq:sub}), the ratio granularity of Eq.~(\ref{eq:seqratio}), and the
training pool differ between runs.
\paragraph{What each variant name denotes.} The names in Tables~\ref{tab:rl} and~\ref{tab:transfer} are
run labels; each is one setting of Eqs.~(\ref{eq:joint}), (\ref{eq:loo}) and~(\ref{eq:sub}) crossed with the ratio of
Eq.~(\ref{eq:seqratio}) and the normalisation of Eq.~(\ref{eq:adv}).
Three labels deserve a word. \textbf{GRPO} is joint credit under the token-level ratio,
that is $\lambda{=}0$ in Eq.~(\ref{eq:loo}); $\lambda$ is the only knob separating it from
removal credit, so those two rows are one estimator at two settings. \textbf{CF} abbreviates
counterfactual and denotes the same removal term as \eqref{eq:loo}. \textbf{Leave-one-out} names leaving one speaker out, and should not be
confused with the leave-one-out baseline over rollouts that appears inside (6);
the latter averages counterfactual substitutions with the other $N-1$ speakers held fixed,
whereas an RLOO baseline would average independent rollouts and give every speaker the same
scalar. Table~\ref{tab:transfer} reuses two of these configurations unchanged under the same names:
its leave-one-out rows are removal credit under the token-level ratio and its GSPO rows
are joint credit under the sequence-level ratio, both retrained on InternVL3.5-8B.

\begin{table}[t]
  \centering\footnotesize
  \setlength{\tabcolsep}{4pt}
  \begin{tabular}{@{}lllcc@{}}
    \toprule
    variant & credit & ratio & $\sigma_R$ & pool \\
    \midrule
    GRPO                     & joint \eqref{eq:joint} & token    & yes & B+C \\
    GSPO                       & joint \eqref{eq:joint} & sequence & yes & B+C \\
    Dr.\ GRPO                  & joint \eqref{eq:joint} & token    & no  & B+C \\
    Leave-one-out (LOO)        & removal \eqref{eq:loo} & token    & yes & B+C \\
    GSPO\,$+$\,LOO             & removal \eqref{eq:loo} & sequence & yes & B+C \\
    Dr.\ GRPO\,$+$\,CF         & removal \eqref{eq:loo} & token    & no  & B+C \\
    C3 substitution            & subst.\ \eqref{eq:sub} & token    & yes & B+C \\
    Leave-one-out (pool ABC)   & removal \eqref{eq:loo} & token    & yes & A+B+C \\
    \bottomrule
  \end{tabular}
  \caption{Every run in Table~\ref{tab:rl} as a setting of the estimators in
  Section~\ref{sec:task:rl}. Seven cross credit with optimiser; the last repeats removal credit on
  the wider pool.}
  \label{tab:supp:variants}
\end{table}

\begin{table}[h]
  \centering\small
  \setlength{\tabcolsep}{5pt}
  \begin{tabular}{@{}lcc@{}}
    \toprule
    variant & B\,$+$\,C & words/msg \\
    \midrule
    centralised bar            & 38.7 & --- \\
    untrained bar              & 28.7 & 43.5 \\
    \midrule
    GRPO                     & 34.7 & 48.6 \\
    GSPO                       & 32.2 & 50.1 \\
    Dr.\ GRPO                  & 27.6 & 138.8 \\
    Leave-one-out (LOO)        & \best{35.2} & 108.0 \\
    GSPO\,$+$\,LOO             & 34.0 & 52.4 \\
    Dr.\ GRPO\,$+$\,CF         & 33.8 & 133.9 \\
    C3 substitution~\cite{chen2026exact} & 21.9 & 131.7 \\
    Leave-one-out (pool A+B+C) & 27.2 & 29.3 \\
    \bottomrule
  \end{tabular}
  \caption{Message length for every variant of Table~\ref{tab:rl}, omitted from the main text
  for space. The budget is $60$ words. The strongest variant on B\,$+$\,C is also the
  second most verbose, while joint credit and the sequence-level objectives stay inside it.}
  \label{tab:supp:lengths}
\end{table}

\cref{tab:supp:lengths} gives the message-length column omitted from Table~\ref{tab:rl}. The
trainer applies no hard length penalty, so nothing prevents a policy from exceeding the stated
budget, and four of the eight variants do so by a wide margin. This matters for how
Table~\ref{tab:rl} should be read. Removal credit leads on B+C at $35.2$ but transmits $108$
words per message, so among the variants that respect the $60$-word budget the leader is
joint credit at $34.7$ with $48.6$ words, ahead of GSPO\,$+$\,LOO at $34.0$ with $52.4$
words and GSPO at $32.2$ with $50.1$ words. 

\paragraph{Reward density.} Eq.~(\ref{eq:reward}) pays a speaker the probability the listener
assigns to the gold option. Replacing that with a binary signal, one when the listener's
top choice is correct and zero otherwise, changes only the density of the reward: the
estimator, the pool, the budget, the group size and the number of updates are unchanged. We
trained this variant on the same pool and for the same number of blocks as the joint-credit
run of Table~\ref{tab:rl} and scored both policies on the full test set under both listeners.

\begin{table}[h]
  \centering\small
  \setlength{\tabcolsep}{6pt}
  \begin{tabular}{@{}lcc@{}}
    \toprule
    reward & CoT listener & scoring listener \\
    \midrule
    listener probability & 34.7 & 30.7 \\
    binary, top-$1$ correct                & 32.9 & 32.9 \\
    \bottomrule
  \end{tabular}
  \caption{Reward density, B\,$+$\,C accuracy (\%). Same backbone, pool, budget and number
  of updates; only the reward changes. Both policies scored on the full test set under both
  listeners.}
  \label{tab:supp:binreward}
\end{table}

The dense reward is worth $1.8$ points under the chain-of-thought listener, which is the
listener Table~\ref{tab:rl} reports. Under the scoring listener the ordering reverses by $2.2$
points. We do not read it as a general claim
about binary rewards and future works should invest time in RL setup designs to improve upon this.

\section{Transfer across backbone and certifier}
\label{sec:supp:transfer}

\begin{table}[h]
  \centering\small
  \setlength{\tabcolsep}{5.5pt}
  \begin{tabular}{@{}ll cc@{}}
    \toprule
    & & \multicolumn{2}{c}{evaluated on} \\
    \cmidrule(lr){3-4}
    method & train pool & Qwen-cert. & IVL-cert. \\
    \midrule
    \multicolumn{4}{@{}l}{\emph{(a) joint credit, scoring listener}} \\
    centralised & ---        & 37.1 & 31.2 \\
    untrained   & ---        & 31.3 & 28.8 \\
    joint       & Qwen-cert. & 30.7 & 29.5 \\
    joint       & IVL-cert.  & \cellcolor{hirow}\best{36.8} & \cellcolor{hirow}\best{31.8} \\
    \midrule
    \multicolumn{4}{@{}l}{\emph{(b) chain-of-thought listener}} \\
    centralised & ---        & 35.9 & 31.9 \\
    untrained   & ---        & 32.2 & 28.3 \\
    removal     & Qwen-cert. & 31.5 & \cellcolor{hirow}\best{32.5} \\
    removal     & IVL-cert.  & 30.3 & 29.1 \\
    GSPO        & Qwen-cert. & 35.3 & 30.4 \\
    GSPO        & IVL-cert.  & \cellcolor{hirow}\best{35.6} & 31.1 \\
    \bottomrule
  \end{tabular}
  \caption{Two additions to Table~\ref{tab:transfer}, B\,$+$\,C accuracy (\%), InternVL3.5-8B.
  Panel (a) adds the third credit scheme under the scoring listener of the main table;
  panel (b) re-reads the main table's two schemes under the chain-of-thought listener.
  Cells already in Table~\ref{tab:transfer} are not repeated.}
  \label{tab:supp:transfercot}
\end{table}

\cref{tab:supp:transfercot} does two things Table~\ref{tab:transfer} does not. Panel~(a) adds the
third credit scheme of Table~\ref{tab:rl} to the transfer study, scored by the same scoring
listener the main table reports, which is also the listener Eq.~(\ref{eq:reward}) defines.
Panel~(b) re-reads the two schemes carried in the main table under the chain-of-thought
listener, which the policies never saw in training and which therefore acts as a held-out
decoder.

One number belongs here because it differs from the pool size quoted elsewhere. Scoring on
both test sets requires each training pool to exclude every scene appearing in either one,
which leaves $4{,}979$ questions for the Qwen-certified pool and $5{,}012$ for the
IVL-certified pool, against the $5{,}541$ of the single-set pool used in Table~\ref{tab:rl}. The
transfer numbers are therefore obtained from about ten percent less training data.
Panel~(a) is the sharper result. Joint credit trained on the IVL-certified pool supplies the
best cell in either column, reaching $36.8$ against a centralised $37.1$ on the
Qwen-certified test set and $31.8$ against a centralised $31.2$ on the IVL-certified one,
where it is level with the centralised reference. The same scheme trained on the
Qwen-certified pool reaches only $30.7$, below the untrained bar of $31.3$. Changing the training pool alone therefore moves this scheme by $6.1$ points,
more than any difference between credit schemes in the table, which makes the pool the
larger of the two factors on this backbone.
Read against Table~\ref{tab:rl}, the ordering does not carry over. On Qwen3-VL-8B removal credit
leads and joint credit follows it; here joint credit leads and removal credit trails. We
therefore make no claim that any credit scheme is portably best, and read Table~\ref{tab:rl} as
specific to the backbone it was measured on.
Panel~(b) shows what survives a decoder that the policies were never trained against. GSPO still
supplies the best Qwen-certified cell at $35.6$, and training on the IVL-certified pool
still transfers to the Qwen-certified test set without penalty. What does not survive is the
magnitude of removal credit, which falls below the untrained bar of $32.2$ on the
Qwen-certified set, at $31.5$ and $30.3$, so its gains are concentrated in the condition
that matches training.

\begin{table}[b]
  \centering\footnotesize
  \setlength{\tabcolsep}{3.2pt}
  \renewcommand{\arraystretch}{1.0}
  \begin{tabular}{@{}cl ccccc@{}}
    \toprule
    & & ALL & A & B & C & B\,$+$\,C \\
    \midrule
    \multirow{4}{*}{\rotatebox[origin=c]{90}{\scriptsize\textsc{bars}}}
      & centralised (CoT)   & 42.6 & 51.7 & 47.8 & 34.8 & 38.7 \\
      & centralised (score) & 39.4 & 49.3 & 51.7 & 28.0 & 35.2 \\
      & untrained (score)   & 36.2 & 52.7 & 42.0 & 23.5 & 29.1 \\
      & untrained (CoT)     & 35.4 & 51.0 & 35.1 & 25.9 & 28.7 \\
    \midrule
    \multirow{2}{*}{\rotatebox[origin=c]{90}{\scriptsize\textsc{dec.}}}
      & \textsc{justify}       & 35.9 & 50.3 & 36.1 & 26.9 & 29.7 \\
      & \textsc{abstain}       & 27.9 & 58.9 & 21.5 & 11.4 & 14.5 \\
    \midrule
    \multirow{5}{*}{\rotatebox[origin=c]{90}{\scriptsize\textsc{topology}}}
      & \textsc{chain}         & 38.0 & 55.5 & 40.0 & 26.3 & 30.4 \\
      & \textsc{ring}          & 37.7 & 54.8 & 46.8 & 23.1 & 30.3 \\
      & \textsc{regular}       & 36.1 & 51.0 & 46.8 & 22.3 & 29.7 \\
      & \textsc{dense}         & 36.5 & 54.5 & 46.3 & 21.2 & 28.8 \\
      & \textsc{hier}          & 35.5 & 53.4 & 42.0 & 21.6 & 27.8 \\
    \midrule
    \multirow{2}{*}{\rotatebox[origin=c]{90}{\scriptsize\textsc{rnds}}}
      & \textsc{adaptive}      & 37.4 & 54.8 & 46.3 & 22.7 & 29.8 \\
      & \textsc{dense4}        & 35.8 & 53.4 & 42.4 & 22.0 & 28.2 \\
    \midrule
    \multirow{3}{*}{\rotatebox[origin=c]{90}{\scriptsize\textsc{init.}}}
      & \textsc{p2p}           & 38.6 & 55.8 & 45.9 & 24.8 & \best{31.2} \\
      & \textsc{ask2}          & 36.9 & 53.4 & 45.9 & 22.9 & 29.8 \\
      & \textsc{query}         & 35.1 & 50.0 & 41.0 & 23.3 & 28.7 \\
    \midrule
    \multirow{4}{*}{\rotatebox[origin=c]{90}{\scriptsize\textsc{content}}}
      & \textsc{schema}        & 36.3 & 50.3 & 44.4 & 24.2 & 30.3 \\
      & \textsc{roles}         & 34.5 & 48.0 & 43.4 & 22.3 & 28.7 \\
      & \textsc{cce}           & 35.4 & 51.4 & 44.9 & 21.4 & 28.5 \\
      & \textsc{slots}         & 35.1 & 52.4 & 42.0 & 21.4 & 27.6 \\
    \midrule
    \multirow{5}{*}{\rotatebox[origin=c]{90}{\scriptsize\textsc{bandwidth}}}
      & \textsc{words128}$^\dagger$ & 38.8 & 54.8 & 48.8 & 24.6 & 31.9 \\
      & \textsc{total\_budget} & 37.6 & 52.4 & 43.9 & 25.6 & 31.2 \\
      & \textsc{bid}           & 38.0 & 54.1 & 45.9 & 24.6 & 31.0 \\
      & \textsc{words32}       & 36.3 & 53.4 & 42.0 & 23.3 & 29.0 \\
      & \textsc{words16}       & 36.6 & 54.8 & 42.0 & 23.1 & 28.8 \\
    \midrule
    \multirow{2}{*}{\rotatebox[origin=c]{90}{\scriptsize\textsc{tools}}}
      & \textsc{zoom}          & 36.8 & 50.7 & 44.9 & 24.8 & 30.9 \\
      & \textsc{ground}        & 37.2 & 52.7 & 43.4 & 24.8 & 30.4 \\
    \midrule
    \multirow{4}{*}{\rotatebox[origin=c]{90}{\scriptsize\textsc{aggreg.}}}
      & \textsc{agg\_hetero}$^\ddagger$ & 40.4 & 60.3 & 50.2 & 23.7 & 31.8 \\
      & \textsc{agg\_cot}      & 38.0 & 54.1 & 39.0 & 27.5 & 31.0 \\
      & \textsc{trunc32}       & 38.1 & 57.2 & 45.9 & 22.9 & 29.8 \\
      & \textsc{agg\_selfcons} & 36.4 & 53.1 & 40.5 & 24.4 & 29.2 \\
    \bottomrule
  \end{tabular}
  \caption{All $27$ inference-only designs on the $969$ test questions, the full version of
  Table~\ref{tab:designs}. The $23$ speaker-side designs use the scoring listener and read
  against the score bar of $29.1$; the four aggregation designs change the listener and
  read against the CoT bar of $28.7$. Seventeen beat their own bar and ten fall below it.
  $^\dagger$\textsc{words128} spends more than twice the $B{=}60$ budget and is not
  admissible. $^\ddagger$\textsc{agg\_hetero} aggregates with a different and larger model,
  so it is not a within-model comparison.}
  \label{tab:supp:designs}
\end{table}

\section{Inference-only designs}
\label{sec:supp:designs}
Results for all 27 design patterns omitted from the main paper due to space constraints are presented in Table \ref{tab:supp:designs}.

\section{Supervised fine-tuning of the speaker}
\label{sec:supp:sft}

\begin{table}[h]
  \centering\small
  \setlength{\tabcolsep}{4pt}
  \begin{tabular}{@{}l ccccc c@{}}
    \toprule
    system & ALL & A & B & C & B\,$+$\,C & w/msg \\
    \midrule
    centralised (CoT)   & 42.6 & 51.7 & 47.8 & 34.8 & 38.7 & --- \\
    untrained (CoT)     & 35.4 & 51.0 & 35.1 & 25.9 & 28.7 & 43.5 \\
    untrained (score)   & 36.2 & 52.7 & 42.0 & 23.5 & 29.1 & --- \\
    \midrule
    SFT 1 epoch, CoT    & 36.7 & 50.3 & 42.4 & 25.9 & 30.9 & 51.9 \\
    SFT 2 epoch, CoT    & 37.7 & 53.1 & 39.0 & 27.5 & 31.0 & 57.2 \\
    SFT 1 epoch, score  & 36.0 & 54.1 & 42.0 & 22.3 & 28.2 & 51.9 \\
    SFT 2 epoch, score  & 38.3 & 57.5 & 46.3 & 22.9 & 30.0 & 57.2 \\
    \midrule
    removal RL, CoT     & 38.9 & 47.6 & 45.9 & 30.5 & \best{35.2} & 108.0 \\
    \bottomrule
  \end{tabular}
  \caption{Supervised fine-tuning of the speaker, as an alternative to reinforcement
  learning on the same B+C pool. Accuracy (\%).}
  \label{tab:supp:sft}
\end{table}

An obvious alternative to reinforcement learning is to clone successful transcripts. We
sample $K{=}4$ message sets per question with the untrained speaker, keep those whose
listener answer was correct, and fine-tune on them, using the same B+C training pool of
$5{,}541$ questions that the reinforcement-learning runs use.
\cref{tab:supp:sft} shows this works, but modestly. Supervised fine-tuning gains $2.2$ to
$2.3$ points on B+C under the chain-of-thought listener against the untrained bar,
roughly a fifth of the gap, where removal credit gains $6.5$. It is, however,
budget-respecting at $51.9$ and $57.2$ words per message, where removal credit needs
$108$, so per word it is the cheaper intervention. The more informative result is where the gain comes from. One epoch moves stratum B from
$35.1$ to $42.4$ and leaves stratum C at $25.9$, unchanged to three significant figures.
The explanation is the filter. A question contributes training pairs only if at least one
of its four samples was already correct, and that condition is not stratum-neutral:
retention is $68.8\%$ on B but $49.9\%$ on C, so the half of C that the untrained policy
never solves is silently dropped. Cloning successes can only teach what the policy already
does occasionally, and on the synergistic stratum that is exactly what it fails at.
RL has no such filter, since every sampled group contributes gradient
through its advantage including the groups that are entirely wrong, which is a mechanistic
reason to expect it to move C where supervision does not.

\section{Free-riding}
\label{sec:supp:freeride}
Holding the algorithm fixed and changing only the training pool from B+C to A+B+C produces
the clearest diagnostic in the paper, recorded in the last row of Table~\ref{tab:rl}. The
resulting policy attains the highest A-stratum accuracy of any variant, $55.1$, while
C-stratum accuracy falls to $24.2$ and B+C to $27.2$, indistinguishable from no training at
all, and mean message length collapses from $108$ words to $29.3$.

The interpretation is direct. Stratum A questions are answerable without any view, so a
policy rewarded on them discovers that terse, uninformative messages suffice, the
listener's priors carrying the answer while describing one's view is wasted effort. The
speakers stop contributing, free-ride on the listener, and the behaviour generalises to
questions where contribution was essential. This follows directly from admitting questions
whose reward is obtainable without communication, and certifying the pool removes it.

\section{Qualitative analysis}
\label{sec:supp:qual}
The full listener reasoning traces for Figure~\ref{fig:qual} run to $832$ words untrained and
$540$ words trained, which is why the figure shows excerpts. We shall release
transcripts for all evaluated configurations.

\paragraph{A second failure mode:} Section~\ref{sec:results:qual} names a failure
that sits alongside false attribution. Speakers describe one object as though it were
several, and the blind listener, with no way to establish that two descriptions denote the
same thing, enumerates the descriptions rather than the objects. Counting questions expose
it cleanly because the error shows up as an integer.
The case below is from CityCube. The
question asks how many tennis courts appear in total across the five aerial views, with
options $1$, $2$, $3$, $4$ and more than $4$. The ground truth is $2$: the same two courts
are visible from three of the viewpoints. \cref{tab:supp:corr} shows what each speaker
reported and how the listener accumulated it.
What makes the case instructive is that the listener sets itself exactly the right task and still fails to accomplish it.

\begin{table}[h]
  \centering\footnotesize
  \setlength{\tabcolsep}{5pt}
  \begin{tabular}{@{}llc@{}}
    \toprule
    speaker & courts reported in its own view & tally \\
    \midrule
    $u_1$ & none visible                        & $0$ \\
    $u_2$ & none visible                        & $0$ \\
    $u_3$ & one, ``bright blue'', upper right   & $+1$ \\
    $u_4$ & one, ``red'', upper left            & $+1$ \\
    $u_5$ & two, ``green'', upper portion       & $+2$ \\
    \midrule
    \multicolumn{2}{@{}l}{listener's answer, at probability $1.00$} & $4$ \\
    \multicolumn{2}{@{}l}{ground truth}                             & \best{$2$} \\
    \bottomrule
  \end{tabular}
  \caption{A correspondence failure. Every message is
  individually accurate and within budget. The same two courts are reported by three
  speakers under three different colours and three view-relative positions, and the blind
  listener adds them up.}
  \label{tab:supp:corr}
\end{table}

\begin{promptbox}[pbList]{Blind listener, reasoning (excerpt)}
\pbt{... we must reconcile the reports by identifying\\
which observers are describing the same scene or\\
overlapping features ... and avoid double-counting.\\
~\\
Step 1: Identify all mentions of tennis courts.\\
 - Observer 3: "A bright blue tennis court is\\
~~~visible in the upper right corner."\\
 - Observer 4: "A red tennis court is visible in\\
~~~the upper left quadrant."\\
 - Observer 5: "Two green tennis courts are\\
~~~visible, adjacent to a large brick building."\\
~\\
... Observers 3, 4, and 5 all mention tennis courts,\\
and all describe a central church with a spire\\
(though color varies) ...}
\end{promptbox}

The listener notices that the church is described with a
steeple by $u_3$, a green spire by $u_4$ and a teal spire by $u_5$, treats the colour
disagreement as a property of viewpoint, and merges the three reports into one church. It
then declines to apply the same reasoning one sentence later, keeping a blue court, a red
court and two green courts as four distinct objects. Nothing in the messages tells it which
choice is correct, because nothing in the messages carries identity: colour varies with
angle and illumination, and ``upper right'' and ``upper left'' are each relative to a
different camera. A sighted agent resolves the reports onto two surfaces by looking. A blind
listener can only guess, and here it guessed consistently for the church and inconsistently
for the courts.
Three properties make this more than an anecdote. The failure is not perceptual, since every
message describes its own view correctly. It is not a budget artefact, since all five
speakers stay inside the $60$-word limit and spend words on the courts. And it is not
detectable from the output, since the listener answers with probability $1.00$: an
unresolvable premise surfaces as a confident answer rather than as expressed uncertainty,
which is why the mode is invisible without ground truth. The same question is answered the
same way in three independent runs, so this is a property of the configuration rather than
of one sample. The remedy it points to is not a larger budget, which the speakers do not
exhaust, but a shared way to refer to an object across views.


\end{document}